\documentclass{article}
\PassOptionsToPackage{numbers}{natbib}

\usepackage[final]{neurips_2025}

\usepackage[utf8]{inputenc} 
\usepackage[T1]{fontenc}    
\usepackage{hyperref}       
\usepackage{url}            
\usepackage{xurl}           
\usepackage{booktabs}       
\usepackage{amsfonts}       
\usepackage{nicefrac}       
\usepackage{microtype}      
\usepackage{xcolor}         
\usepackage{array}
\usepackage{booktabs} 
\usepackage{makecell} 
\usepackage{multirow} 
\usepackage{graphicx}  
\usepackage{graphicx}
\usepackage{caption}
\usepackage{float}
\usepackage{lipsum}
\usepackage{multirow}
\usepackage{booktabs}
\usepackage{amsmath} 
\usepackage{minitoc}
\usepackage{multicol}
\usepackage{epigraph}
\usepackage{wrapfig}
\usepackage{tcolorbox}
\tcbuselibrary{listingsutf8} 
\usepackage{multirow}
\usepackage{graphicx}
\usepackage{amssymb, amsmath, amsthm}
\usepackage{bbm}
\usepackage{mathtools}
\usepackage{amsfonts}
\usepackage{bm}
\usepackage{array}
\usepackage{xurl}
\usepackage{comment}
\usepackage{enumitem}
\usepackage{graphicx} 
\usepackage{booktabs}         
\usepackage{threeparttable}   

\usepackage{amsmath,amsfonts,bm}

\def\eqref#1{equation~\ref{#1}}

\def\1{\bm{1}}

\def\vw{{\bm{w}}}
\def\vx{{\bm{x}}}
\def\vy{{\bm{y}}}
\def\vz{{\bm{z}}}

\DeclareMathAlphabet{\mathsfit}{\encodingdefault}{\sfdefault}{m}{sl}
\SetMathAlphabet{\mathsfit}{bold}{\encodingdefault}{\sfdefault}{bx}{n}

\usepackage{authblk}      
\usepackage{hyperref}     
\usepackage{fontawesome5} 

\usepackage{authblk}

\newcommand\blfootnote[1]{%
  \begingroup
  \renewcommand\thefootnote{}\footnote{#1}%
  \addtocounter{footnote}{-1}%
  \endgroup
}

\definecolor{deepblue}{RGB}{0,81,102} 
\definecolor{darkblue}{rgb}{0, 0, 0.5}
\newcommand{\revision}[1]{\textcolor{black}{#1}}

\newcommand{\ours}{\textsc{InterMT}}
\newcommand{\oureval}{\textsc{InterMT-Bench}}
\newcommand{\oursettingone}{\textit{interleaved multimodal understanding and generation}}
\newcommand{\oursettingtwo}{\textit{multi-turn}}
\definecolor{myblue}{rgb}{0.488, 0.865, 0.975}
\definecolor{myred}{RGB}{139,0,0} 
\definecolor{lightgray}{rgb}{0.9,0.9,0.9}
\definecolor{multiturn}{RGB}{139,0,0}
\definecolor{understanding}{RGB}{47, 85, 151}
\definecolor{generation}{RGB}{84, 130, 53}

\newcommand{\multiturn}[1]{\textbf{\textcolor{multiturn}{#1}}}
\newcommand{\understanding}[1]{\textbf{\textcolor{understanding}{#1}}}
\newcommand{\generation}[1]{\textbf{\textcolor{generation}{#1}}}

\hypersetup{
    colorlinks=true,
    linkcolor=red,
    citecolor=blue,
    filecolor=magenta,      
    urlcolor=blue,
linktocpage}

\title{\ours{}: Multi-Turn Interleaved Preference Alignment with Human Feedback}

\author[1,2]{Boyuan Chen$^{*}$}
\author[1,2]{Donghai Hong$^{*}$}
\author[1,2]{Jiaming Ji$^{*}$}
\author[3]{Jiacheng Zheng}
\author[1]{Bowen Dong}
\author[1,2]{Jiayi Zhou}
\author[1]{Kaile Wang}
\author[1]{Juntao Dai}
\author[1]{Xuyao Wang}
\author[1]{Wenqi Chen}
\author[1]{Qirui Zheng}
\author[1]{Wenxin Li}
\author[3]{Sirui Han}
\author[3]{Yike Guo}
\author[1]{Yaodong Yang$^{\dag}$}

\affil[1]{Institute for AI, Peking University}
\affil[2]{State Key Laboratory of General Artificial Intelligence, Peking University}
\affil[3]{Hong Kong University of Science and Technology}

\affil[ ]{
    \vspace{0.5em} 
    \texttt{\{cbylll, donghai.hong, jiamg.ji\}@stu.pku.edu.cn} \\
    \texttt{yaodong.yang@pku.edu.cn}
}

\begin{document}
\maketitle
\blfootnote{$^*$ Equal contribution, $^\dag$ Corresponding author. Project website: \url{https://pku-intermt.github.io.}}

\vspace{-3.5em}
\begin{abstract}
As multimodal large models (MLLMs) continue to advance across challenging tasks, a key question emerges: \textbf{\textit{What essential capabilities are still missing? }}
A critical aspect of human learning is continuous interaction with the environment -- not limited to language, but also involving multimodal understanding and generation.
To move closer to human-level intelligence, models must similarly support \textbf{multi-turn}, \textbf{multimodal interaction}. In particular, they should comprehend interleaved multimodal contexts and respond coherently in ongoing exchanges.
In this work, we present \textbf{an initial exploration} through the \textsc{InterMT} -- \textbf{the first preference dataset for \textit{multi-turn} multimodal interaction}, grounded in real human feedback. In this exploration, we particularly emphasize the importance of human oversight, introducing expert annotations to guide the process, motivated by the fact that current MLLMs lack such complex interactive capabilities. \textsc{InterMT} captures human preferences at both global and local levels into nine sub-dimensions, consists of 15.6k prompts, 52.6k multi-turn dialogue instances, and 32.4k human-labeled preference pairs. 
To compensate for the lack of capability for multi-modal understanding and generation, we introduce an agentic workflow that leverages tool-augmented MLLMs to construct multi-turn QA instances.
To further this goal, we introduce \textsc{InterMT-Bench} to assess the ability of
MLLMs in assisting judges with multi-turn, multimodal tasks.
We demonstrate the utility of \textsc{InterMT} through applications such as judge moderation and further reveal the \textit{multi-turn scaling law} of judge model.
We hope the open-source of our data can help facilitate further research on aligning current MLLMs to the next step.
\end{abstract}

\section{Introduction}

Humans perceive the world through dynamic, multimodal interactions involving text, images, audio, video, and more \citep{liu2024improved,turk2014multimodal,zhang2023llama, zhang2025safevla}. Building on the success multimodal large language models (MLLMs) \citep{liu2024llava, wu2024nextgpt, zhao2023survey, yin2023survey,ji2023ai}, recent efforts aim to develop general-purpose AI assistants that handle multiple mixed modalities \citep{alayrac2022flamingo,liu2023visual,team2024chameleon}. A key feature of such general-purpose assistants is to engage in natural \oursettingtwo{} conversations, perceive and generate any modality (\textit{i.e.}, \oursettingone{}), to enable more smooth interaction and grounded understanding \citep{alayrac2022flamingo,liu2024visual,team2024chameleon,yu2024rlhf,ji2024align,min2023factscore}.

Recent years have seen community efforts in transplanting alignment techniques (\textit{e.g.}, Reinforcement Learning from Human Feedback (RLHF)) from the text modality \citep{bai2022training, ji2023ai, ji2024aligner, rafailov2024direct, zhou2025sequence, ji-etal-2025-language-models, meng2025med} to multiple modalities settings \citep{yu2024rlhf,yu2024rlaif,sun2023aligning, Maaz2023VideoChatGPT,majumder2024tango,wallace2024diffusion,ji2024align, zhou2025generative, ji2025safe}. Within this line of research, most studies focus exclusively on either understanding \citep{zhang2025mm,yu2024rlaif} or generation \citep{wallace2024diffusion,majumder2024tango}. The lack of alignment considerations for multimodal mixed input-output settings exacerbates the imbalance across modalities, \textit{i.e.}, \textit{modality disequilibrium}.\citep{ji2024align}. Furthermore, existing methods primarily focus on single-turn interactions, where an LLM generates a response from a prompt and receives immediate alignment feedback.
However, real-world interactions typically occur in long-horizon conversations (\textit{e.g.}, over 5 turns) and often feature interleaved multimodal inputs and outputs \citep{liu2024holistic,he2024multi,sirdeshmukh2025multichallenge}.

\begin{tcolorbox}[top=1pt, bottom=1pt, left=0.7pt, right=0.7pt, boxsep=2pt, colframe=white]
\begin{center}
\small
\textit{How to improve \multiturn{multi-turn} interleaved \understanding{understanding}- \generation{generation} alignment via human feedback?}
\end{center}
\end{tcolorbox}

Our reflections highlight several key issues in the alignment of MLLMs:
\begin{itemize}[left=0.3em]
    \item \textbf{Modality Fusion via Harmonizing Understanding and Generation.}  
    To build general-purpose AI assistants, high-fidelity perception and understanding alone are not sufficient. The system should also support the selective generation of multimodal outputs (\textit{e.g.}, images) to effectively communicate, instruct, or interact with users in a natural and contextually appropriate manner.
    \item \textbf{Modeling Long‐Horizon, Interleaved Multimodal Interactions.}  
    Real‐world user–AI exchanges typically span many turns and interleave text, vision, and other modalities. Such interactions demand not only precise instruction following but also sustained attention and reasoning over an evolving context, approaching near‐human in‐context reasoning capabilities.
    \item \textbf{Dynamic Human‐in‐the‐Loop Alignment.}  
    In extended, multimodal interactions, user preferences continually evolve. For example, a user may first ask the assistant to draw a vase, then—after inspecting the rendered image—request that the vase be repositioned or restyled for greater emphasis. Capturing and aligning with these emergent, dynamic preferences calls for genuine, iterative human feedback throughout the interaction.
\end{itemize}

\begin{figure}
\centering
\includegraphics[width=\linewidth]{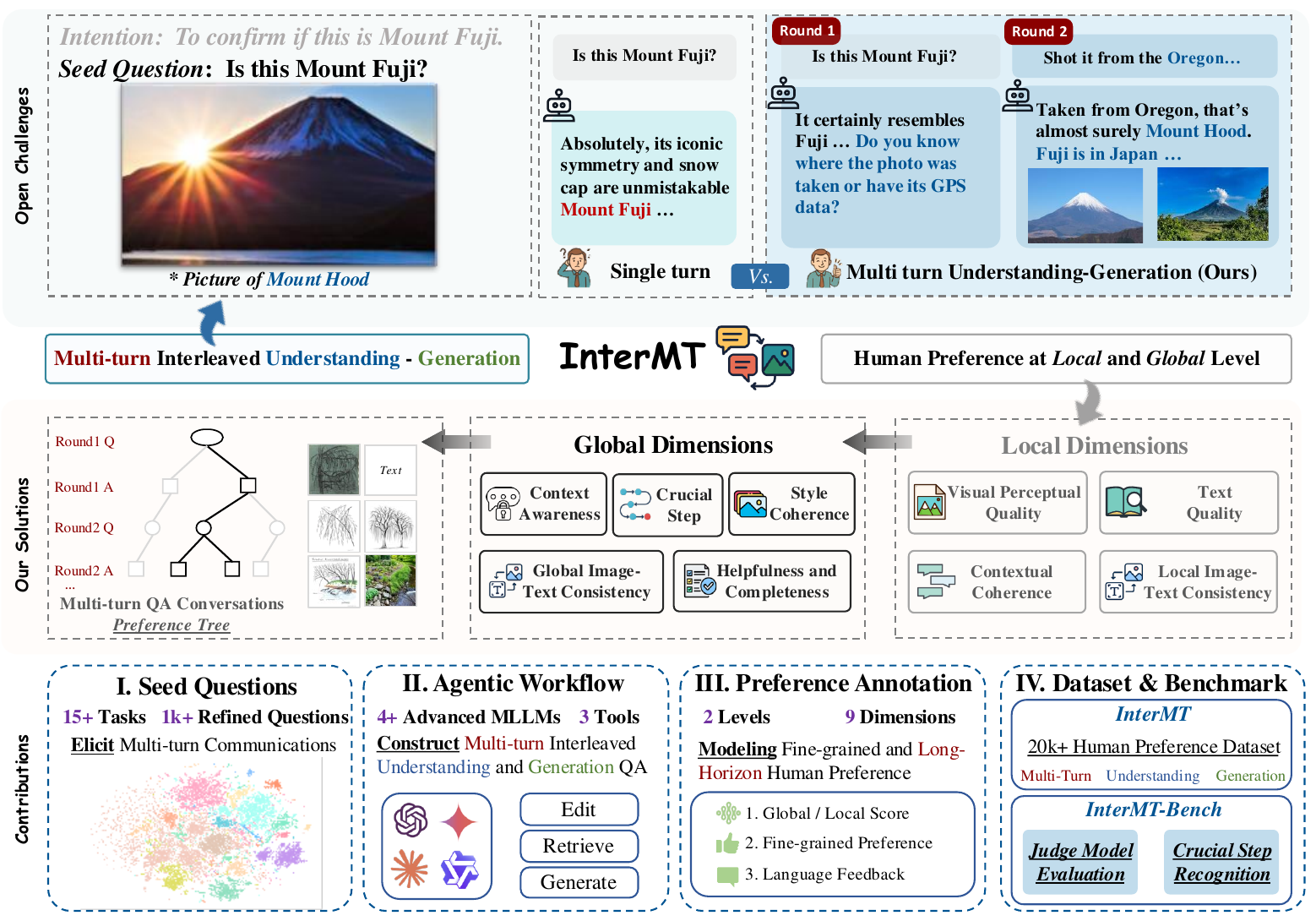}
\vspace{-1.5em}
\caption{Motivated by the challenges of single-turn interactions in aligning with human intent, and the goal of constructing general-purpose AI assistants, we introduce \textbf{\ours{}}: 
\textbf{Data:} \ours{}, the first human preference dataset focused on \multiturn{multi-turn}, multimodal \understanding{understanding} and \generation{generation}; 
\textbf{Decoupled Helpfulness:} capturing human feedback at both the \textit{local} (turn-level) and \textit{global} (conversation-level); 
\textbf{Evaluation:} evaluating the capabilities of MLLMs as judge models.}
\label{fig:overall}
\end{figure}

In response, we introduce \ours{}, a human preference dataset designed to capture the complexity and diversity of human intent in \multiturn{multi-turn} settings. Specifically, \ours{} targets vision-language interaction scenarios involving interleaved \understanding{understanding} and \generation{generation}. To model dynamic human preferences, \ours{} comprises 15604 seed questions that elicit multi-turn, multimodal conversations spanning 15+ domains. Helpfulness is then decomposed into 9 sub-dimensions, capturing both global (conversation-level) and local (turn-level) aspects of human feedback.

Our key contributions are summarized as follows:
\begin{itemize}[left=0.3em]
    \item \textbf{The First Multi-turn Interleaved Preference Dataset:} To the best of our knowledge, \ours{} is the first dataset that captures real human preferences for tasks involving \multiturn{multi-turn} and \textit{interleaved multimodal \understanding{understanding} and \generation{generation}}. It contains 15604 unique seed questions across diverse categories, 52.6k multi-turn interleaved vision-language QA instances, and 32,459 sets of multi-dimensional human preference annotations.
    \item \textbf{Agent-based Construction Workflow:} \ours{} employs a carefully designed agent-based multi-turn QA construction workflow that leverages strong MLLMs augmented with external tools (\textit{e.g.}, image editing, generation and retrieval) to simulate high-quality real multi-turn interactions.
    \item \textbf{Decoupled Helpfulness in Multi-turn Multimodal Scenarios:} \ours{} decomposes the concept of helpfulness for \oursettingtwo{} \oursettingone{} into two distinct levels: \textit{local} (turn-level) and \textit{global} (conversation-level). At the local level, helpfulness is assessed for each individual turn, while at the global level, helpfulness is evaluated across the entire conversation. Furthermore, \ours{} breaks down helpfulness into 9 specific dimensions (\textit{e.g.}, \textit{contextual coherence}, \textit{image-text consistency}, etc.), allowing for a detailed and nuanced evaluation of multi-turn, multi-modal interactions.
    \item \textbf{Effective for Multi-turn Alignment:} Building on \ours{}, we investigate methods to \textit{model long-horizon values} and \textit{align dynamic human values}. Our findings reveal the phenomenon of preference transfer in multi-turn multimodal interactions, which facilitates preference modeling for predicting human judgments. Additionally, we identify a \textit{scaling phenomenon} in multi-turn multimodal judge moderation (Section \ref{sec:prefernece_modeling_multiturn}). 
    \item \textbf{One More Thing} We introduce \oureval{} to evaluate the ability of MLLMs in assisting judges across multi-turn, multimodal tasks, encompassing three parts: \textit{Scoring Evaluation}, \textit{Pair Comparison}, and \textit{Crucial Step Recognition} (Section \ref{sec:oureval}). Despite strong reasoning capabilities, advanced MLLMs (\textit{e.g.}, o4-mini \citep{openai2025o4mini}) fail to align with human values in judgment tasks. However, they show potential in identifying crucial steps in long-context scenarios.

\end{itemize}
For more details about the motivation of our work, please refer to Appendix \ref{app:related_work}.

\section{Dataset}

Our core contribution is the introduction of a human preference dataset designed for \multiturn{multi-turn}, multimodal \understanding{understanding} and \generation{generation} tasks. This section outlines the dataset's composition, the collection of prompts and multi-turn QA instances, and human annotation process.

\begin{figure}[t]
    \centering
    \includegraphics[width=\columnwidth]{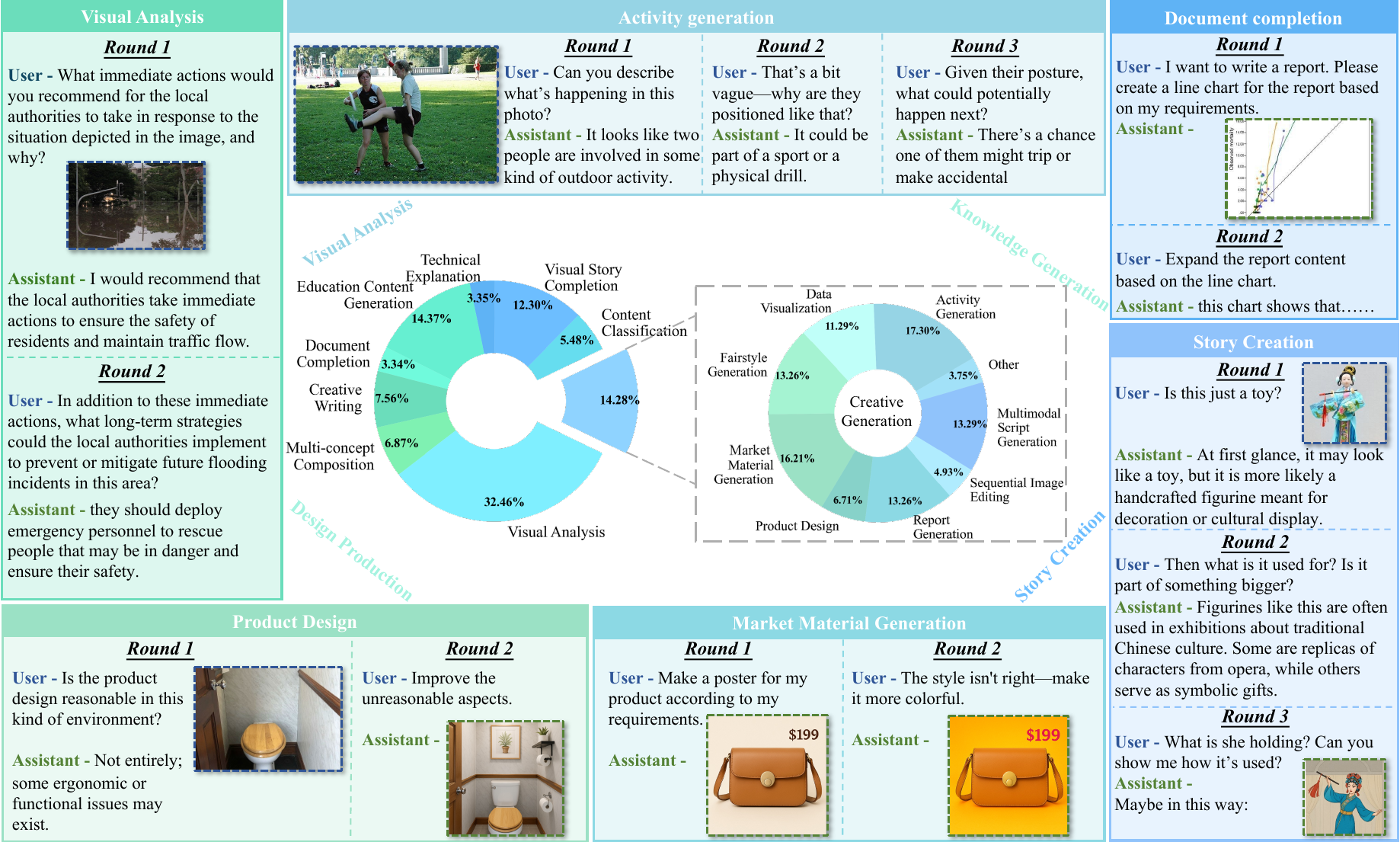}
    \caption{\ours{} includes over 15 tasks in vision-language scenarios, capturing communication examples across diverse \multiturn{multi-turn} settings. These examples demonstrate \multiturn{multi-turn}, interleaved \understanding{understanding} and \generation{generation} in six representative domains.}
    \vspace{-0.5em}
    \label{fig:seed_prompt-category}
\end{figure}

\subsection{Dataset Composition}

The \ours{} dataset includes: (1) carefully crafted \textit{seed questions} for multi-turn, multimodal conversations, and (2) fine-grained human preference annotations at both local and global conversation levels. Inspired by theories from linguistics, human-computer interaction, and cognitive psychology \citep{grice1975logic,grosz1995centering,clark1991grounding,parsing2009speech,traum1995computational,pku2025deception}, the seed questions are rigorously selected and refined to enable more faithful simulation of real-world \oursettingone{} and \oursettingtwo{} tasks.
We collect preference data through score evaluations and pairwise comparisons of multi-modal responses at each conversation turn, based on four sub-dimensions. Global conversation helpfulness is then evaluated via five sub-dimensions. Incorporating natural language feedback further improves annotation quality and alignment with human intent.
The \textbf{Data Card} for \ours{} is as follow:
\begin{itemize}[left=0.3em]
    \item \ours{} is built from a corpus of 100k image-text examples, comprising 72.1\% from open-source vision-language datasets, 22.8\% from web data, and 5.1\% from human-written content. All prompts are refined following constitutional guidelines to improve multi-turn compatibility, resulting in 15604 unique seed questions, as shown in Figure \ref{fig:seed_prompt-category}.
    \item Each seed question is expanded via an agent-based multi-turn QA construction workflow, producing at least 8 multi-turn QA instances per prompt. After pruning and filtering, we obtain 52.6k high-quality multi-turn QA instances, with 41.92\% containing five or more turns.
    \item The resulting 52.6k QA instances cover 15+ vision-language \understanding{understanding} and \generation{generation} tasks, such as image editing and visual tutorials. Each instance features interleaved textual and visual content in both inputs and outputs, with an average of 5.33 images per conversation.
    \item \ours{} features 32,459 human preference annotations, organized as score evaluation pairwise comparisons at both the local and global levels. Preferences are decomposed into 9 dimensions of helpfulness, accompanied by human-written critiques, refinement suggestions, and rationales.
\end{itemize}

\subsection{Multi-turn QA Construction}
\begin{figure}[t]
    \includegraphics[width=\columnwidth]{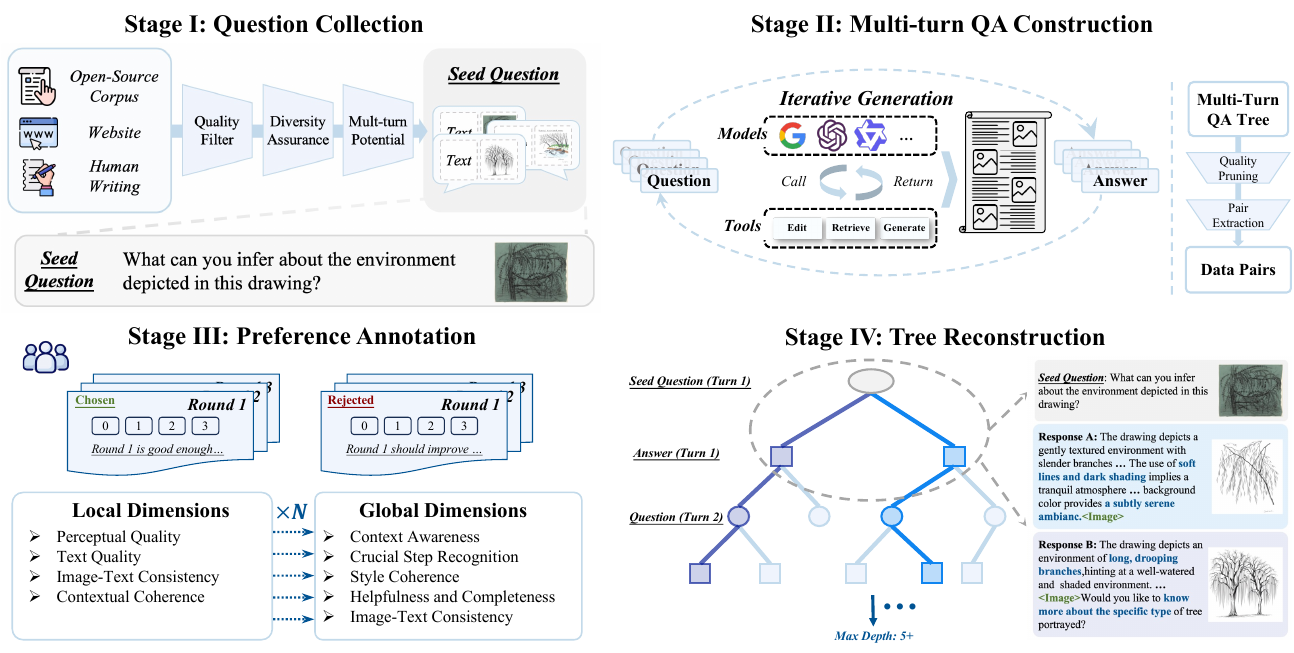}
    \vspace{-1.5em}
    \caption{Overview of the four‐stage pipeline for constructing \ours{}. \textbf{Stage I:} seed questions are harvested from open‐source corpora, websites, and human writing, then filtered for perceptual quality, diversity, and multi‐turn potential. \textbf{Stage II:} iterative calls to large models and external tools (\textit{e.g.} edit, retrieve, generate) produce answer expansions and follow-up questions, forming a candidate QA tree. \textbf{Stage III:} human annotators perform per‐turn (local) and conversation-level (global) evaluations—covering quality, coherence, context awareness, and completeness—to prune and select preferred branches. \textbf{Stage IV:} the retained branches are reassembled into deep, coherent QA trees (depth $\geq$ 5) yielding the final multi‐turn QA pairs for model training.}
    \vspace{-0.5em}
    \label{fig:multi_turn_qa_construction_framework}
\end{figure}

\noindent \textbf{Prompt Collection.~}
\label{sec:prompt_collection}
\ours{} is constructed from 100k image-text QA instances collected from three primary sources: 72.1\% from public datasets \citep{ji2024align,zhang2025mm,chen2024comm}; 22.8\% from legally scraped web content; and the remaining 5.1\% from researcher-curated, human-written prompts. These instances span diverse vision-language tasks, \textit{e.g.}, activity generation, data visualization, and table analysis.

Drawing upon cognitive psychology theories \citep{grice1975logic,grosz1995centering,clark1991grounding,parsing2009speech,traum1995computational}, we identify five common scenarios that give rise to multi-turn conversations in real-world multimodal settings. Based on these scenarios, we filter, diversify, and rewrite the original image-text QA instances, resulting in 15604 unique \textit{seed question}s. These questions serve as the initial round for generating multi-turn conversation data. Additional details can be found in Appendix \ref{app:data_collection}.

\noindent \textbf{Tool-Augmented Agent Workflow for QA Construction.~}
\label{sec:tool_augmented_agent_workflow_for_qa_construction}
We identify two core challenges in constructing multi-turn QA instances that capture realistic scenarios of multimodal understanding and generation: (1) How to effectively simulate realistic human multi-turn conversations in multimodal contexts? (2) Given that current MLLMs lack interleaved understanding and generation capabilities \citep{xia2025mmie,liu2024holistic}, how to construct interleaved QA instances that generalize across diverse real-world tasks?

To address these challenges, we propose a tool-augmented agent workflow that integrates powerful open-source and API-based models with image-centric tools. Within this framework, each agent simulates human-like conversations by either responding to the current query or generating follow-up questions based on the previous answer. Agents can invoke tools to generate, edit, or retrieve images, enabling the recursive construction of tree-structured, multi-turn interleaved image–text QA instances.

\noindent \textbf{Agent Construction.~} 
The agent workflow is built upon a combination of strong open-source models \citep{Qwen2VL, bai2025qwen2, team2025gemma, liu2024llava} alongside leading API-based models \citep{openai2024gpt4o, google2025gemini2flash, anthropic2024claude3}.
To support diverse multimodal operations, three types of image-centric tools are integrated: (1) text-to-image generators (\textit{e.g.}, FLUX.1-Schnell \citep{black2024flux} and Stable-Diffusion \citep{rombach2021highresolution}) for producing high-quality images based on prompts; (2) an image editing API (\textit{e.g.}, Gemini-2.0-flash \citep{google2025geminiimagegeneration}) capable of cropping, highlighting, and modifying images; and (3) web-based retrieval interfaces
for sourcing real-world visuals.
During multi-turn QA generation, agents embed structured tokens such as \texttt{<Image,\textit{caption}>} within the text to denote visual references after which GPT-4o \citep{openai2024gpt4o} serves as a double classifier and verifier, automatically determining the appropriate tool call based on the image intent and context.

\noindent \textbf{Iteratively Question and Response Generation.}
We begin with carefully crafted \emph{seed questions} to initiate extended multimodal dialogues; at each turn, diverse agents generate a pool of 10 candidate follow-ups via a Socratic strategy, from which $\mathcal{M}$ (typically 1–3) high-quality, non-redundant questions are selected using textual similarity ranking and regex filtering, ensuring contextual coherence and, when needed, visual clarification. Each selected follow-up is then answered by sampling over 10 candidate responses paired with multiple visual options, from which $\mathcal{N}$ (typically 2–4) responses are chosen based on relevance and multimodal quality, with optional user-guided continuations to enhance satisfaction. Repeating this selection process for $n$ rounds yields a tree-structured QA dataset of size $\prod_{i=1}^{n}\mathcal{M}_i\times\mathcal{N}_i\,$. For more details, see Appendix \ref{app:data_collection}.

\noindent \textbf{Quality Control and Pruning.~}
We apply a filtering strategy from multiple perspectives with two key components: the image(-text) Filter, which evaluates each candidate image for visual quality and semantic relevance, and the consistency filter, which preserves content and stylistic coherence across dialogue turns. Finally, we prune the multi-turn paths based on overall quality, coherence, and diversity, yielding a refined set of QA instances for annotation.

\noindent \textbf{Human Annotation.~}
Defining high-quality multi-turn multimodal dialogues is inherently challenging, as it requires assessing response correctness, the coherence of image-text interleaving, and the dynamic nature of human preferences throughout the conversation. We conduct multiple rounds of in-depth discussions with our annotation team regarding existing open-source datasets and prior work on MLLMs.
We then identify the following 9 annotation dimensions.

\begin{multicols}{2}
\begin{itemize}[left=0cm]
\setlength\itemsep{-0.25em}
 \item G1: Context Awareness
 \item G2: Helpfulenss and Completeness
 \item G3: Crucial Step Recognition
 \item G4: Global Image-Text Consistency
 \item G5: Style Coherence
 \item L1: Local Image-Text Consistency
 \item L2: Visual Perceptual Quality
 \item L3: Contextual Coherence 
 \item L4: Text Quality
\end{itemize}
\end{multicols}

Crowdworkers first rate individual turns and then evaluate entire conversations from both local and global perspectives. A \textbf{Dual Verification} stage combines dedicated annotator efforts with professional quality control reviews to ensure guideline adherence. Structured \textbf{Language Feedback}, which offers concise explanations of scoring rationale, focused critiques, and refinement suggestions, further guides response improvement and substantially enhances annotation reliability.

\section{Analysis}
\label{sec:analysis}

Since the \ours{} dataset captures \textit{real} human preferences across multiple dimensions at both \textit{global} and \textit{local} levels, it is meaningful to analyze the correlations among these dimensions, examine the relationship between per-turn preferences and overall evaluation, and further compare human feedback with AI feedback in this section.

\begin{figure}[htbp]
    \includegraphics[width=\columnwidth]{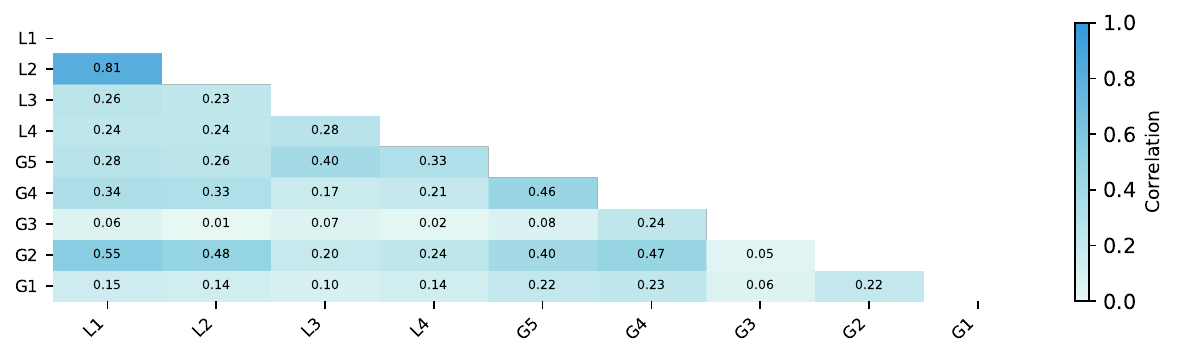}
    \vspace{-2em}
    \caption{Linear correlation coefficient of different preference annotations. }
    \vspace{-0.5em}
     \label{fig:data_statistics_analysis}
\end{figure}

\begin{wrapfigure}[16]{r}{0.5\linewidth}
    \vspace{-20pt}
    \centering
    \includegraphics[width=\linewidth]{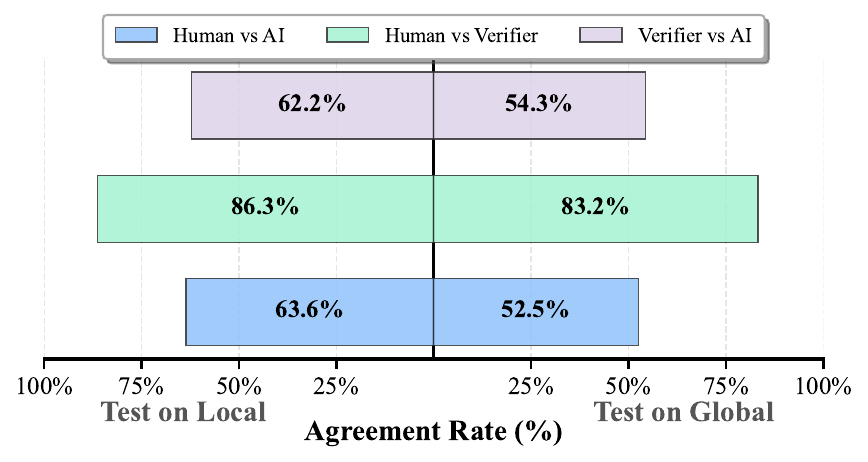}

\caption{Agreement rates between different roles at local and global levels. Here, \textit{Human} refers to the outsourced annotation team, while \textit{Verifier} denotes human experts with professional backgrounds and the authors, who perform random sampling and manually verify 5–10\% of the dataset.}

    \label{fig:human_ai_agreement_analysis}
\end{wrapfigure}

\noindent \textbf{Correlation Analysis.~}
Figure \ref{fig:data_statistics_analysis} illustrates the relationship between global and local preference annotation dimensions. 
We identify three key findings:
\textbf{(1) Modality perception precedes effective modality fusion}: for both the local-local and local-global correlation, the evaluation of image-text consistency is strongly correlated with visual perceptual quality (up to 0.81). This suggests that before assessing multimodal information, human evaluators tend to prioritize a clear understanding of each individual modality, indicating that a clear perception of individual modalities is a prerequisite for reliable multimodal judgment.
\textbf{(2) Long-horizon evaluations hinge on coherence and temporal consistency}: for the global-global correlation, metrics such as helpfulness and completeness strongly align with context awareness and global visual consistency, underscoring the importance of maintaining coherent semantics, multimodal information, and consistency with prior conversational context over extended interactions.
\textbf{(3) Intent grounding drives long-horizon crucial step recognition}: in multi-turn scenarios, models may deviate from the user’s core intentions, producing self-directed responses. Despite locally high-scoring and plausible outputs, this leads to stylistic drift and omission of key steps over extended interactions, as demonstrated in the local-global correlation setting.

\noindent \textbf{Human Feedback \textit{vs.} AI Feedback.}
Human-labeled data introduce high cost, which motivates the exploration of MLLMs' potential to assist with evaluation tasks \citep{chen2024mllm}. We develop a pipeline that utilizes advanced API-based models \citep{openai2024gpt4o, openai2025gpt41, google2025geminipro, google2025gemini2flash, anthropic2025claude37, openai2025o4mini}) to produce multidimensional scores from both global and local perspectives. Then, we evaluate the agreement between AI and human annotators, as well as between AI annotators and expert human verifiers. Agreement scores are then averaged across all pairs for comparative analysis. Experimental results (Figure~\ref{fig:human_ai_agreement_analysis}) show that while AI annotators achieve approximately 60\% agreement on local evaluations, their consensus with humans on global (longer‐horizon) tasks is markedly lower. This indicates current MLLMs struggle to match human judgments in multi‐turn, multimodal scoring. 
Until AI feedback efficacy is firmly established, replacing human annotation remains inadvisable.

\section{Inspiring Future Research}
\label{sec:inspiring_future_research}

\ours{} lays the groundwork for advancing research on aligning human values in \textit{multi-turn multimodal understanding and generation} tasks, potentially inspiring new research directions. Building on real human data provided by \ours{}, we identify several promising directions:

\begin{itemize}[left=0.3em]
\item \textbf{Modeling long-horizon values.} How can we model long-horizon, interleaved multimodal preferences by leveraging the \textit{local} and \textit{global} human annotations in \ours{}?
\item \textbf{Aligning dynamic human values}: How can we design algorithms that effectively incorporate real human feedback from \ours{} to assess and enhance the performance of MLLMs?
\end{itemize}

In this section, we present several baseline approaches that address the above questions, with the goal of fostering further research and demonstrating the utility of our dataset.

\subsection{Preference Modeling for Multi-turn Interleaved Multimodal Scenarios}
\label{sec:prefernece_modeling_multiturn}
A widely adopted approach for modeling human preferences is to employ a preference predictor grounded in the Bradley–Terry (BT) model \citep{bradley1952rank}. 
However, when extending to multi-turn settings, new challenges arise—particularly in capturing the dynamics of evolving user preferences across turns. Moreover, traditional outcome-level reward signals often fail to generalize in purely textual domains \citep{shani2024multi}, let alone in complex multimodal settings involving interleaved understanding and generation. 
\ours{} incorporates both \textit{local} and \textit{global} human annotations in multi-turn, multimodal interactions, leading us to investigate efficient preference modeling methods.

\begin{wrapfigure}[14]{r}{0.5\linewidth}
    \vspace{-16pt}
    \centering
    \includegraphics[width=\linewidth]{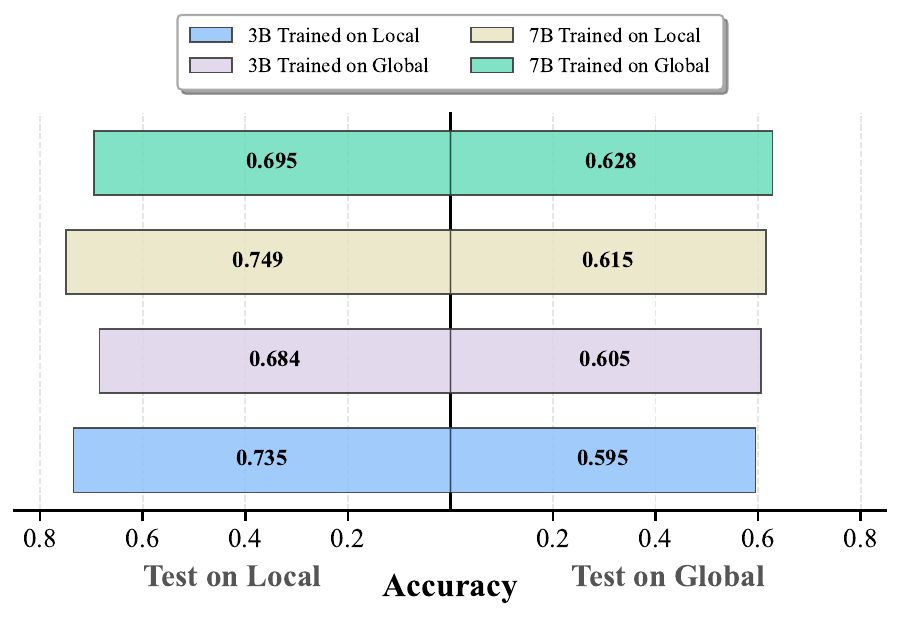}
    \vspace{-2.0em}
    \caption{Judge models trained and evaluated on different dataset.}
    \label{fig:local-global-compare}
\end{wrapfigure}

Inspired by \citep{qiu2024reward,liao2025tpo, zhang2012human}, we investigate two strategies for modeling long-horizon preferences in multi-turn multimodal scenarios: \textit{prefix preference} and \textit{chain-based preference}. Details of formulations can be seen in Appendix \ref{app:experiment_details}.
 Our findings, presented in Figure~\ref{fig:local-global-compare}, suggest that modeling fine-grained \textit{local (turn-level)} preferences is more effective in capturing human values and achieving better alignment. In contrast, directly modeling \textit{global (conversation-level)} preferences often fails to reflect these nuanced preferences, especially in complex, long-horizon scenarios.

\noindent\textbf{Local \textit{vs.} Global Preference Transfer.~}
We examine the bidirectional transfer between turn-level (\textit{local}) and conversation-level (\textit{global}) human preferences. As shown in Figure \ref{fig:local-global-compare}, both \textit{local-to-global} and \textit{global-to-local} transfers are effective, since multi-turn questions typically hinge on the seed question’s intent. However, \textit{global-to-local} transfer is consistently easier and better aligned with actual preferences. We attribute this to the greater stability of global preferences—reflecting users’ overarching tendencies—whereas local preferences are short-term and more context-dependent, making \textit{local-to-global} transfer more challenging.

\paragraph{Multi-turn Scaling Law of Turn-based Judge Moderation}
\textit{Can we accurately capture users' intentions and latent preferences with a limited number of conversational turns, thereby improving the modeling of long-term values?} Such capabilities are crucial for building general-purpose AI assistants, which need to understand and predict users' needs across diverse contexts, adapting to changing preferences over time. We investigate whether the discriminative power of judge models, trained on the first \( k \) turns, improves in subsequent turns (from \( k+1 \) to \( N \)) and exhibits \textit{scaling laws}. 

\begin{figure}[t] 
        \vspace{0pt}
        \centering
        \includegraphics[width=\linewidth]{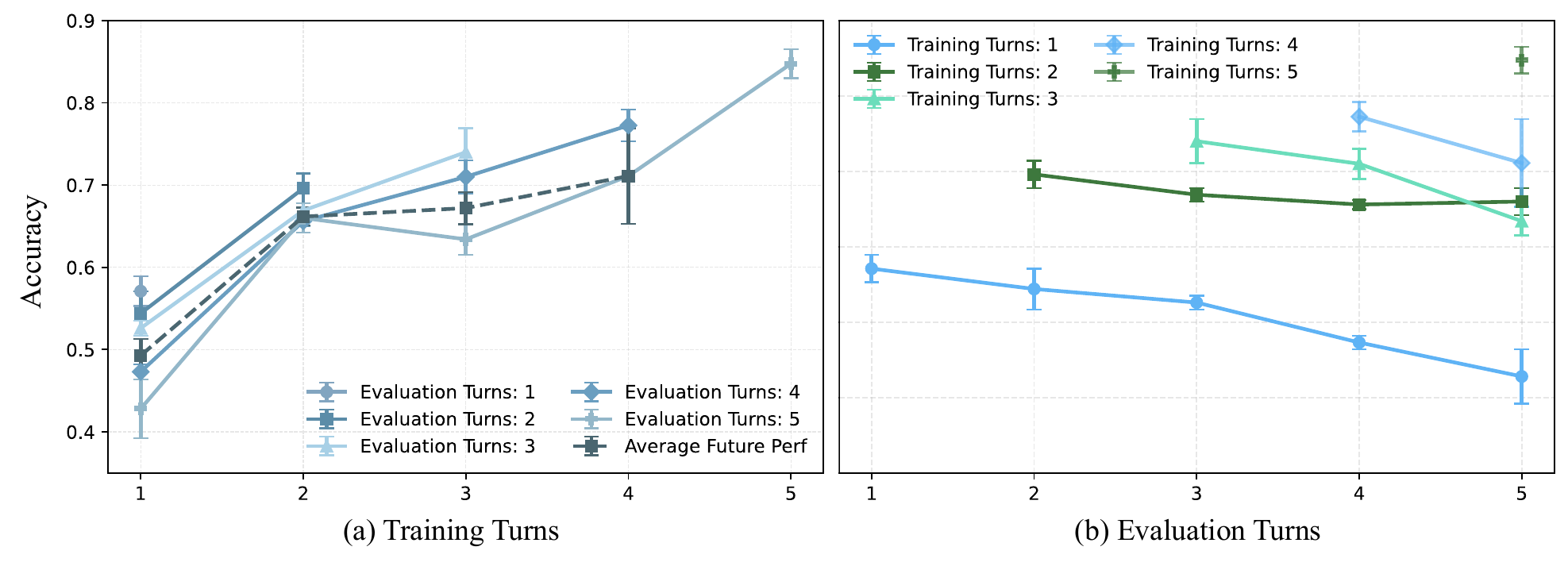}
        \vspace{-1.5em}
        \captionof{figure}{Scaling laws of judge models. As training turns increase, model’s ability to predict future preferences improves (left), while generalization diminishes as evaluation turns increases (right). } 
        \label{fig:multi_turn_scaling_laws_of_judge_model}
\end{figure}

The results reveal two key insights: (1) Multi-turn judge moderation exhibits a generalization effect linked to the number of turns. As shown in Figure \ref{fig:multi_turn_scaling_laws_of_judge_model} (a), for evaluation turn $k$, as the number of preceding turns increases from 1 to $k-1$, the model's accuracy continues to improve, with average future performance rising, indicating that training on multi-turn data with a limited number of turns can generalize to longer horizons. (2) Regarding the number of turns in the training data, the generalization effect shows a diminishing trend. As demonstrated in Figure \ref{fig:multi_turn_scaling_laws_of_judge_model} (b), training with $k$ turns does improve performance for $k+1 \rightarrow T$ turns, but this effect diminishes as the number of turns increases. The decline is due to three factors: diminishing returns as the model struggles with long-term preferences, contextual drift as earlier turns lose relevance, and the evolving interaction between user intentions and latent preferences.

\subsection{MLLM as a Judge and \oureval{}}
\label{sec:oureval}

\textit{Do MLLMs truly understand what is desirable in multi-turn, multimodal interactions and how to align with human values?} This task is particularly challenging due to the absence of multimodal benchmarks that capture human preferences in multi-turn settings. Inspired by \citep{zheng2023judging,chen2024mllm} and leveraging genuine feedback from \ours{}, we introduce \oureval{} to assess MLLMs' alignment with human values in multi-turn, multimodal tasks.
\oureval{} comprises three distinct tasks: \textit{Scoring Evaluation}, \textit{Pair Comparison}, and \textit{Crucial Step Recognition}. Details can be found in Appendix \ref{app:details_of_oureval}.

\paragraph{Judge Settings and Metrics} The dataset includes multi-turn multimodal interleaved communication histories and human-annotated ground truth. Evaluated models must assess the conversation at both the turn and conversation levels across nine dimensions, following a set of guidelines. \textit{Scoring Evaluation} requires the model to assign scores on a 0-3 scale, with evaluation based on Pearson similarity \citep{lee1988thirteen,zheng2023judging,chen2024mllm}. \textit{Pair Comparison} directly compares two individual turns or entire conversations, without considering ties, and is evaluated for accuracy against human judgments. \textit{Crucial Step Recognition} test if MLLMs can accurately identify the user’s intent and determine whether it has been fulfilled, evaluated by the score provided by judge according to the human-annotated reference answers.

\begin{figure}[t] 
        \vspace{0pt}
        \centering
        \includegraphics[width=\linewidth]{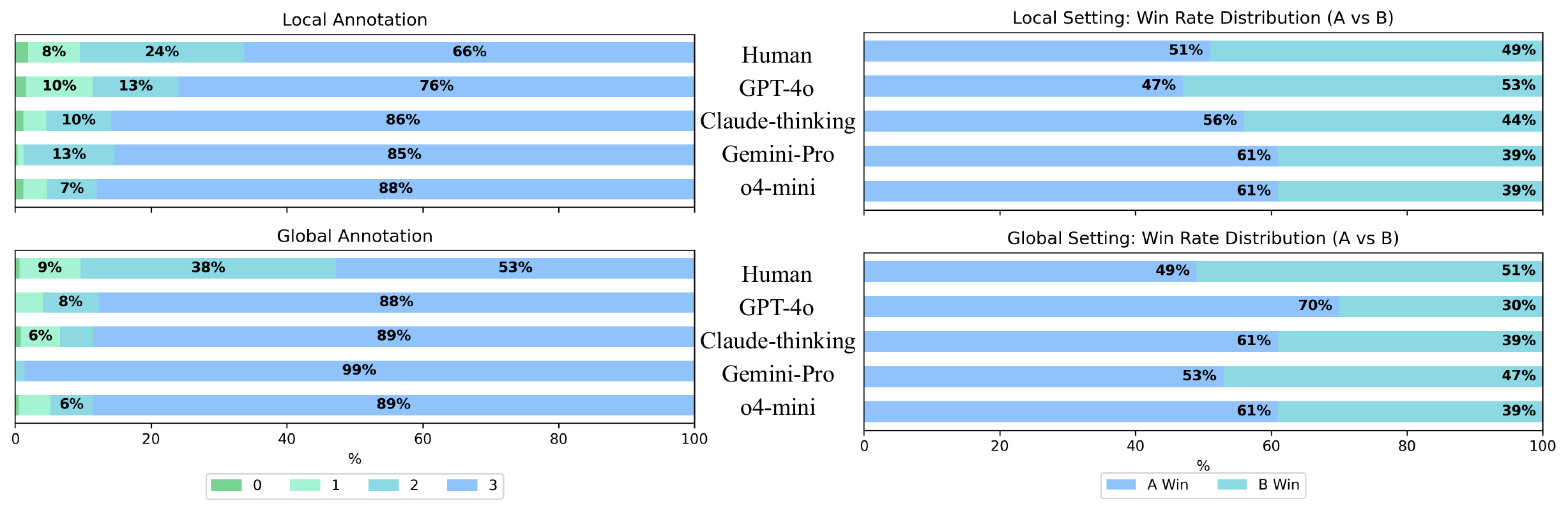}
        \vspace{-1.5em}
        \captionof{figure}{Score and length distribution comparison.}
        \vspace{-0.5em}
        \label{fig:distribution_bias}
\end{figure} 

\paragraph{Results and Takeaways} We evaluated 6 advanced MLLMs for their ability to assist in judgment for multi-turn multimodal interactions, considering the nine dimensions proposed above. The results reveal key observations:  \textbf{Existing models still face challenges in aligning with long-horizon human values, but they perform more accurately in evaluating local, fine-grained preferences}. As shown in Table \ref{tab:our_eval_results}, all models exhibit significant gaps in performance compared to humans in both Score Evaluation and Pair Comparison tasks. However, the models demonstrate better accuracy when assessing local dimensions rather than global dimensions, suggesting that capturing fine-grained (\textit{e.g.}, turn-level) human preferences is crucial for both evaluation and alignment with human dynamic and long-horizon values. 
However, there is cause for optimism: current MLLMs exhibit near-human-level performance (4.38/5) in recognizing task completion and aligning with human intent (\textit{i.e.}, \textit{Crucial Step Recognition}), providing potential solutions for long-term value alignment.

\begin{table}[t]
\centering
\caption{Overall performance comparison of various MLLMs across three judgment tasks in \oureval{}. The Score Evaluation task requires models to assign scores on a 0–3 scale and is evaluated using Pearson correlation metric. The Pair Comparison task involves direct binary comparison of two individual turns or entire conversations without ties and is evaluated by accuracy against human judgments. All reported Pearson correlation values achieve statistical significance with p-values below 0.05. \textit{Note:} Gemini-Flash* denotes Gemini-2.0-Flash, Gemini-Pro* corresponds to Gemini-2.5-Pro-preview, and Claude-thinking* refers to the Claude-3.7-Sonnet (thinking) model.}
\resizebox{\textwidth}{!}{%
\begin{tabular}{l|p{4.5cm}|ccccc|cccccc}
\toprule
\multirow{3}{*}{\textbf{Settings}} & \multirow{3}{*}{\textbf{MLLMs}} & \multicolumn{5}{c|}{\textbf{Local Setting}} & \multicolumn{6}{c}{\textbf{Global Setting}} \\
\cmidrule(lr){3-7} \cmidrule(lr){8-13} 
& & \textbf{L1} & \textbf{L2} & \textbf{L3} & \textbf{L4} & \textbf{Avg.} & \textbf{G1} & \textbf{G2} & \textbf{G3} & \textbf{G4} & \textbf{G5} & \textbf{Avg.}\\
\midrule 

\multirow{12}{*}{\textbf{\begin{tabular}[l]{@{}c@{}}Scoring \\ Evaluation\end{tabular}}}
& Gemini-Flash*  & 0.346 & 0.107 & 0.119 & 0.173 & 0.186 & 0.163 & 0.042 & 0.051 & \textbf{0.246} & 0.005 & 0.101 \\
& Gemini-Flash* (+reason) & \textbf{0.361} & 0.072 & 0.122 & 0.168 & 0.181 & -0.038 & 0.083 & 0.139 & 0.199 & 0.048 & 0.086 \\
& GPT-4.1 & 0.264 & 0.095 & 0.242 & 0.269 & 0.218 & 0.215 & 0.216 & 0.084 & 0.044 & 0.049 & 0.122 \\
& GPT-4.1 (+reason) & 0.281 & 0.094 & 0.272 & 0.271 & 0.229 & 0.215 & 0.255 & 0.217 & 0.216 & 0.050 & \textbf{0.191} \\
& GPT-4o & 0.291 & \textbf{0.131} & 0.277 & 0.268 & \textbf{0.242} & 0.254 & 0.167 & 0.137 & 0.139 & 0.069 & 0.153 \\
& GPT-4o (+reason) & 0.290 & 0.091 & 0.252 & \textbf{0.280} & 0.228 & 0.183 & 0.243 & 0.194 & 0.086 & 0.072 & 0.156 \\
& Gemini-Pro* & 0.273 & 0.079 & 0.258 & 0.168 & 0.194 & \textbf{0.285} & 0.240 & -0.024 & 0.235 & \textbf{0.145} & 0.176 \\
& Gemini-Pro* (+reason) & 0.274 & 0.070 & 0.304 & 0.211 & 0.215 & 0.239 & \textbf{0.267} & 0.195 & 0.129 & 0.060 & 0.178 \\
& Claude-thinking* & 0.299 & 0.044 & 0.262 & 0.229 & 0.209 & 0.172 & 0.140 & 0.175 & 0.150 & 0.069 & 0.141 \\
& Claude-thinking* (+reason) & 0.291 & 0.023 & 0.254 & 0.214 & 0.196 & 0.207 & 0.260 & 0.183 & 0.155 & -0.001 & 0.161 \\
& o4-mini & 0.334 & 0.062 & 0.306 & 0.134 & 0.209 & 0.169 & 0.161 & 0.120 & 0.096 & 0.028 & 0.115 \\
& o4-mini (+reason) & 0.326 & 0.056 & \textbf{0.322} & 0.151 & 0.214 & 0.215 & 0.229 & \textbf{0.347} & 0.137 & 0.016 & 0.189 \\
\midrule
\multirow{10}{*}{\textbf{\begin{tabular}[l]{@{}c@{}}Pair \\ Comparison\end{tabular}}}
&GPT-4.1                    & 0.541          & \textbf{0.589} & 0.508          & 0.484          & 0.531          & 0.540          & 0.520          & 0.530          & \textbf{0.590} & \textbf{0.563} & \textbf{0.549} \\
&GPT-4.1 (+reason)          & 0.550          & 0.584          & 0.501          & 0.521          & \textbf{0.539} & 0.520          & 0.520          & 0.477          & 0.513          & 0.540          & 0.514          \\
&GPT-4o                     & 0.513          & 0.488          & 0.499          & 0.510          & 0.503          & 0.560          & 0.517          & \textbf{0.550} & 0.543          & 0.470          & 0.528          \\
&GPT-4o (+reason)           & 0.500          & 0.537          & \textbf{0.511} & 0.509          & 0.514          & 0.542          & 0.490          & 0.545          & 0.522          & 0.528          & 0.525          \\
&Gemini-Pro*                & 0.533          & 0.521          & 0.496          & 0.533          & 0.521          & \textbf{0.562} & \textbf{0.566} & 0.523          & 0.505          & 0.505          & 0.532          \\
&Gemini-Pro*   (+reason)    & 0.526          & 0.528          & 0.513          & 0.514          & 0.520          & 0.548          & 0.562          & 0.495          & 0.522          & 0.538          & 0.533          \\
&Claude-thinking*           & 0.561          & 0.568          & 0.508          & 0.502          & 0.535          & 0.539          & 0.523          & 0.518          & 0.521          & 0.528          & 0.526          \\
&Claude-thinking* (+reason) & \textbf{0.567} & 0.550          & 0.506          & 0.519          & 0.536          & 0.512          & 0.522          & 0.512          & 0.547          & 0.512          & 0.521          \\
&o4-mini                    & 0.556          & 0.549          & 0.508          & \textbf{0.536} & 0.537          & 0.552          & 0.498          & 0.522          & 0.518          & 0.495          & 0.517          \\
&o4-mini (+reason)          & 0.521          & 0.564          & 0.522          & 0.513          & 0.530          & 0.534          & 0.510          & 0.507          & 0.512          & 0.483          & 0.509 \\

\bottomrule 
\end{tabular}
} 
\label{tab:our_eval_results}
\end{table} 


\begin{wrapfigure}{r}{0.5\textwidth}
    \centering
    \vspace{-1.0em}
    \includegraphics[width=0.5\textwidth]{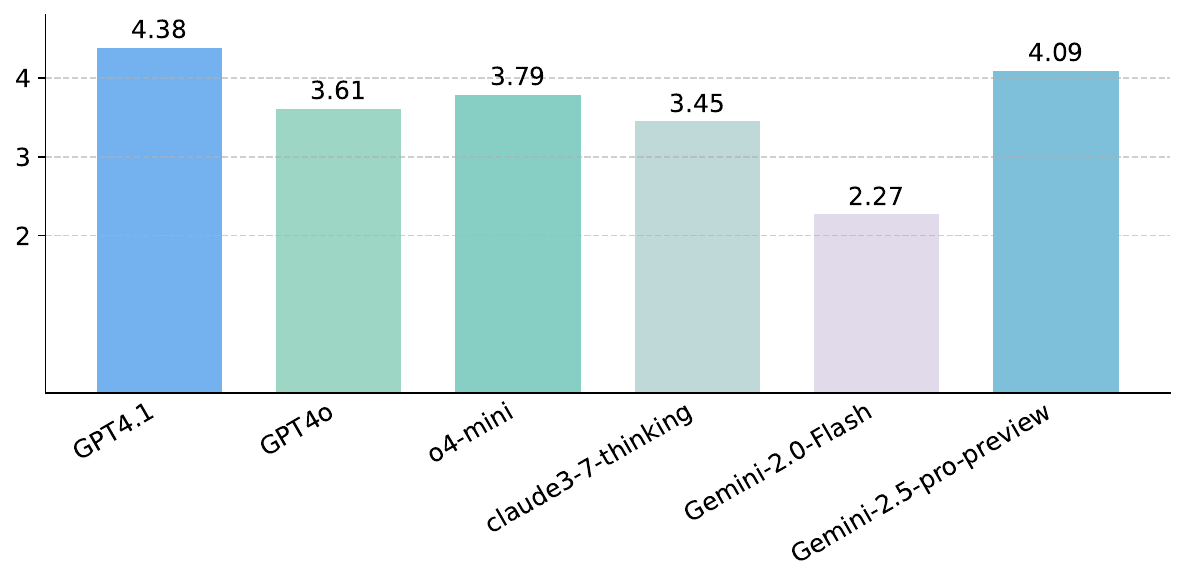}
    \caption{Results of \textit{Crucial Step Recognition}.}
    \vspace{-2.0em}
    \label{fig:crucial_step_recognition_score}
\end{wrapfigure}

\noindent \textbf{Induced Bias and Hallucination.~}
Consistent with \citep{chen2024mllm}, we identify issues related to bias and hallucination: \textbf{Position Bias}, where models consistently favor responses in specific positions (\textit{e.g.}, the first answer), often influenced by training data that places correct answers at the beginning or end of prompts \citep{li2023generative}, and \textbf{High-Score Bias} \citep{chen2024mllm}, where models tend to assign higher scores to entire multi-turn communications. These issues, particularly in long-horizon tasks, may hinder the model’s ability to capture differences between extended conversations, thereby posing challenges in modeling long-horizon values and potentially leading to safety concerns \citep{anil2024many}.

\noindent \textbf{Reasoning Ability Is Not a Panacea.}
\revision{We compare weak reasoning (\textit{i.e.}, providing plausible explanations for the evaluation results) with strong reasoning (\textit{i.e.}, using advanced reasoning models like o4-mini) on the scoring evaluation and pair comparison settings. However, the results are suboptimal. The models' reasoning processes are primarily based on a step-by-step comparison against predefined guidelines, rather than actively identifying potential flaws in the responses. This approach, which differs from the more granular feedback humans provide, leads to misalignments with human preferred judgement.}

\noindent \textbf{Divide and Conquer is beneficial for Crucial Step Recognition} 
\revision{We observe that models with high scores in \textit{Crucial Step Recognition} (shown in Figure \ref{fig:crucial_step_recognition_score}) tend to adopt a divide-and-conquer approach, meaning they first assess each turn in a multi-turn, multimodal dialogue for problem-solving, and then provide an overall conclusion; in contrast, models with lower scores often give more generalized and ambiguous responses.}

\section{Conclusion and Outlook}
\label{sec:conclusion_and_outlook}
This work introduces \ours{}, the first human preference dataset designed for multi-turn, multimodal understanding and generation tasks, capturing human feedback at both local (turn-level) and global (conversation-level) granularities across nine dimensions. We also present \oureval{} to evaluate the capability of advanced MLLMs in assisting with judging such complex interactions. We find that modeling fine-grained local (turn-level) preferences is generally more effective in capturing human values and achieving better alignment compared to directly modeling global (conversation-level) preferences. Analyzing preference transfer, we observe that while both local-to-global and global-to-local transfers are effective, global-to-local transfer is consistently easier and better aligned with actual preferences. Another key observation is the multi-turn scaling law of judge moderation: as the number of training turns increases, the model's ability to predict future preferences improves, while its generalization ability diminishes with longer evaluation horizons. 

\paragraph{Fair and Ethical Labor}
We employed 30 full-time, experienced crowdsourced workers for multimodal annotation, ensuring fair and transparent compensation significantly above the local minimum wage, in compliance with labor regulations. The \ours{} project underwent thorough ethical review by the Institutional Review Board and the Academic Committee of the Institution for Artificial Intelligence at Peking University. The dataset will be released under a CC BY-NC 4.0 license. To mitigate potential risks, NSFW filtering was applied during dataset construction, though we acknowledge that absolute safety cannot be ensured. Given the heightened societal risks associated with multimodal data, we recommend safeguards such as gated access. We are committed to the responsible development of AI and strongly oppose any malicious or unethical use of the \ours{} dataset.

\subsection{Limitations and Future Work}
While our work, utilizing the carefully curated \ours{} dataset, provides significant clarity on modeling long-horizon preferences in vision-language settings, it also illuminates the inherent complexity of human communication and thus reveals important directions for future research. A key observation is that human communication extends far beyond vision and language; it is deeply multimodal, involving video, audio, and more. Future work is essential to extend \ours{} to encompass these additional modalities, moving closer to a holistic representation of communications dynamics. Future work should focus on developing more advanced and efficient methods specifically tailored to exploit the unique richness of the local and global human annotations in \ours{}, thereby enhancing model alignment.
Finally, as MLLMs are increasingly used in diverse real-world contexts, ensuring robust value alignment and safety becomes a critical challenge. Addressing the complexities of diverse human values will require sustained, interdisciplinary research efforts.

\section*{Acknowledgment}
This work is sponsored by the National Natural Science Foundation of China (62376013, 623B2003, 624B100026). Any opinions, findings, conclusions, or recommendations expressed in this material are those of the author(s) and do not necessarily reflect the views of the funding agencies.
\bibliographystyle{unsrt}
\bibliography{main}

\newpage
\section*{NeurIPS Paper Checklist}

\begin{enumerate}

\item {\bf Claims}
    \item[] Question: Do the main claims made in the abstract and introduction accurately reflect the paper's contributions and scope?
    \item[] Answer: \answerYes{} 
    \item[] Justification: We accurately stated our contributions and scope in the abstract and introduction, and provided the open-source link for our dataset.

\item {\bf Limitations}
    \item[] Question: Does the paper discuss the limitations of the work performed by the authors?
    \item[] Answer: \answerYes{} 
    \item[] Justification: As shown in Section \ref{sec:conclusion_and_outlook}, we discuss our limitations and future directions.

\item {\bf Theory assumptions and proofs}
    \item[] Question: For each theoretical result, does the paper provide the full set of assumptions and a complete (and correct) proof?
    \item[] Answer: \answerNA{} 
    \item[] Justification: This paper does not contains theoretical results that should be proved.

    \item {\bf Experimental result reproducibility}
    \item[] Question: Does the paper fully disclose all the information needed to reproduce the main experimental results of the paper to the extent that it affects the main claims and/or conclusions of the paper (regardless of whether the code and data are provided or not)?
    \item[] Answer: \answerYes{} 
    \item[] Justification: We provide our detailed experiment results in the Appendix and we also open-source all of our code and data.

\item {\bf Open access to data and code}
    \item[] Question: Does the paper provide open access to the data and code, with sufficient instructions to faithfully reproduce the main experimental results, as described in supplemental material?
    \item[] Answer: \answerYes{} 
    \item[] Justification: We provide all of our code and data, and can be accessed by our project link.

\item {\bf Experimental setting/details}
    \item[] Question: Does the paper specify all the training and test details (e.g., data splits, hyperparameters, how they were chosen, type of optimizer, etc.) necessary to understand the results?
    \item[] Answer: \answerYes{} 
    \item[] Justification: Yes, we clarify necessary details in the main body of paper and give detailed information in the Appendix to help reproduction and understanding.

\item {\bf Experiment statistical significance}
    \item[] Question: Does the paper report error bars suitably and correctly defined or other appropriate information about the statistical significance of the experiments?
    \item[] Answer: \answerYes{} 
    \item[] Justification: Yes, we provide experiment results in several benchmarks and all results are double-verified by human experts. We also provide detailed human agreement analysis in the body and the Appendix of paper.

\item {\bf Experiments compute resources}
    \item[] Question: For each experiment, does the paper provide sufficient information on the computer resources (type of compute workers, memory, time of execution) needed to reproduce the experiments?
    \item[] Answer: \answerYes{} 
    \item[] Justification: We provide detailed information of the computer resources in the Appendix.
    
\item {\bf Code of ethics}
    \item[] Question: Does the research conducted in the paper conform, in every respect, with the NeurIPS Code of Ethics \url{https://neurips.cc/public/EthicsGuidelines}?
    \item[] Answer: \answerYes{} 
    \item[] Justification: We carefully check our paper that conform with the NeurIPS Code of Ethics in every respect.

\item {\bf Broader impacts}
    \item[] Question: Does the paper discuss both potential positive societal impacts and negative societal impacts of the work performed?
    \item[] Answer: \answerYes{} 
    \item[] Justification: Yes, we discuss the fair use and potential societal impacts in Section \ref{sec:conclusion_and_outlook}. 
    
\item {\bf Safeguards}
    \item[] Question: Does the paper describe safeguards that have been put in place for responsible release of data or models that have a high risk for misuse (e.g., pretrained language models, image generators, or scraped datasets)?
    \item[] Answer: \answerYes{} 
    \item[] Justification: Yes, and we discuss them in Section \ref{sec:conclusion_and_outlook}. Given that multimodal data may pose greater societal risks than pure text data, we believe it is necessary to consider implementing safeguards for sensitive content, such as adopting Hugging Face’s gated dataset access settings.

\item {\bf Licenses for existing assets}
    \item[] Question: Are the creators or original owners of assets (e.g., code, data, models), used in the paper, properly credited and are the license and terms of use explicitly mentioned and properly respected?
    \item[] Answer: \answerYes{} 
    \item[] Justification: Yes, our dataset and other resources are released under the \textbf{CC BY-NC 4.0} License. We also discuss the license in the Appendix.

\item {\bf New assets}
    \item[] Question: Are new assets introduced in the paper well documented and is the documentation provided alongside the assets?
    \item[] Answer: \answerYes{} 
    \item[] Justification: Yes, we provide detailed documentation in the Appendix.

\item {\bf Crowdsourcing and research with human subjects}
    \item[] Question: For crowdsourcing experiments and research with human subjects, does the paper include the full text of instructions given to participants and screenshots, if applicable, as well as details about compensation (if any)? 
    \item[] Answer: \answerYes{} 
    \item[] Justification: Yes, we provide detailed documentations in the Appendix.

\item {\bf Institutional review board (IRB) approvals or equivalent for research with human subjects}
    \item[] Question: Does the paper describe potential risks incurred by study participants, whether such risks were disclosed to the subjects, and whether Institutional Review Board (IRB) approvals (or an equivalent approval/review based on the requirements of your country or institution) were obtained?
    \item[] Answer: \answerYes{} 
    \item[] Justification: Yes, we provide IRB in our supplementary materials

\item {\bf Declaration of LLM usage}
    \item[] Question: Does the paper describe the usage of LLMs if it is an important, original, or non-standard component of the core methods in this research? Note that if the LLM is used only for writing, editing, or formatting purposes and does not impact the core methodology, scientific rigorousness, or originality of the research, declaration is not required.
    \item[] Answer: \answerYes{} 
    \item[] Justification: We use LLMs for the help of dataset construction. All of the experiment details and proper use of LLMs are provided and discussed in the Appendix 

\end{enumerate}

\newpage

\doparttoc
\faketableofcontents
\part{Appendix}

\parttoc
\appendix

\clearpage

\section{Related Work}
\label{app:related_work}

\subsection{QA Dataset with Human-Preference Annotation}
Human preference annotations are essential for aligning language models with the 3H objectives: helpfulness, harmlessness, and honesty \citep{askell2021general, bai2022training, ji2023ai}. These preferences are typically converted into reward signals via the Bradley-Terry model \citep{bradley1952rank}, facilitating the use of established RL methods \citep{ouyang2022training} or direct policy optimization toward preferred response distributions \citep{rafailov2024direct}. A number of datasets offer question–answer pairs with human preference annotations, ranging from safety-focused datasets \citep{xu2021bot, ouyang2022training, ji2023beavertails, ji2024pku} to multimodal preference datasets \citep{ji2024align, yu2024rlhf,dai2024safesora,kirstain2023pick,li2024vlfeedback,zhang2024spa}. However, preference data in multi-turn dialogue settings remains underexplored. Existing studies primarily focus on multi-turn response generation \citep{shani2024multi}, rather than improving instruction-following quality across turns, particularly in multimodal contexts. \ours{} fills this gap by introducing a dataset specifically designed for preference-based human annotation dataset in multi-turn, multimodal interactions.

\subsection{Interleaved Image-Text Dataset}

Training on interleaved image-text web documents has shown superior performance compared to simple image-description pairs, as demonstrated by models such as Flamingo \citep{alayrac2022flamingo}, Chameleon \citep{team2024chameleon}, and MiniGPT-5 \citep{zheng2023minigpt}. This improvement is attributed to the richer and more meaningful correlations in interleaved documents, underscoring their importance in developing interleaved generation models. However, the training data used in these studies is not publicly available. Recent efforts have focused on constructing interleaved image-text datasets \citep{feng2022mmdialog,das2017visual,mostafazadeh2017image,shuster2018image,chen2024comm}. For instance, MMC4 \citep{zhu2023multimodal} extends the text-only C4 dataset \citep{raffel2020exploring,googlec4} by incorporating images into text documents. OBELICS \citep{laurencon2023obelics} collects large-scale data from web pages. However, both datasets suffer from low image-text coherence and a limited number of images per document \citep{chen2024comm}. Other datasets focus on image-centered question answering \citep{das2017visual,mostafazadeh2017image,shuster2018image}; however, their limited task diversity and reasoning depth reduce their suitability for high-quality visual instruction tuning. CoMM \citep{chen2024comm}, which sources data from websites such as WikiHow, emphasizes visual tutorials. Nevertheless, none of these datasets support multi-turn interactions.
To address these limitations, we present \ours{}—a multi-turn image-text interaction dataset encompassing diverse tasks, including visual instruction following, image editing, causal reasoning and so on. \ours{} emphasizes image-text coherence, and logical consistency across dialogue turns, aiming to enhance the general instruction-following capabilities of MLLMs.

\subsection{Multi-Turn QA Dataset}
Recent studies have concentrated on constructing multi-turn dialogue datasets, typically through human-human interactions, to facilitate the development of more effective chat-based AI assistants. These datasets generally incorporate both vision and text modalities \citep{mostafazadeh2017image,das2017visual,shuster2018image,feng2022mmdialog}. However, these datasets are restricted to image-text inputs and textual multi-turn outputs, which are often collected via crowdsourcing under narrowly defined tasks. Consequently, the resulting data often contain colloquial expressions, making them suboptimal for enhancing instruction-following capabilities. More importantly, these studies lack a principled methodology for constructing multi-turn preference datasets. To advance any-modality MLLMs, there is still a notable scarcity of high-quality vision-language interactive datasets that incorporate human-annotated preference.

\subsection{Modeling of Interleaved Image-Text}

The advent of multimodal large language models has markedly advanced tasks involving interleaved text-image understanding and generation. Earlier models like DALL-E \citep{ramesh2021zero} and Stable Diffusion \citep{podell2023sdxl} showcased impressive capabilities in generating high-quality images from textual descriptions, whereas models such as LLaVA \citep{liu2023visual, liu2024llava} achieve notable breakthroughs in image understanding and reasoning via vision instruction tuning.
However, previous research has predominantly focused on unidirectional generation—either text-to-image or image-to-text—without addressing interleaved generation scenarios in which text and images are seamlessly integrated within the same input or output. Recent efforts have begun to close this gap \citep{wu2023next,wu2024nextgpt,alayrac2022flamingo,sun2024generative,li2025uni,awadalla2023openflamingo,xie2024show,zhang2025nexus}. Flamingo \citep{alayrac2022flamingo} introduced image tokens into the language modeling process, whereas Chameleon proposed a unified architecture embedding both modalities into a shared space for multimodal input and output. Emu \citep{sun2024generative} utilizes Stable Diffusion \citep{podell2023sdxl} as an image decoder, thereby enabling generation from interleaved image-text inputs.
Despite these advances, existing models continue to struggle with multimodal contextual consistency—such as semantic coherence and stylistic alignment between images and text \citep{wu2024nextgpt, chen2024comm, liu2024holistic, xia2025mmie}. Furthermore, multi-turn dialogue capabilities—such as contextual coherence, modality-aware content selection, and heuristic question generation—remain underexplored. To address these gaps, we present \ours{}—a human preference dataset specifically designed for multi-turn interleaved text-image  understanding and generation.

\subsection{AI Alignment and RLHF}

Aligning LLMs with human preferences is critical for their safe and effective deployment \citep{ji2023ai}. Among various approaches, supervised fine-tuning (SFT) and reinforcement learning from human feedback (RLHF) have emerged as standard methods for aligning model behavior with human intent \citep{ouyang2022training,bai2022training,rafailov2024direct}. Recent work has extended this alignment framework beyond language-only settings to multimodal scenarios involving both image and text modalities \citep{yu2024rlhf,zhang2025mm,yu2024rlaif,ji2024align,chen2025ai}. Such multimodal alignment necessitates addressing challenges like interleaved image-text inputs and outputs, alongside multi-turn interactions that reflect real-world usage. However, the approach to alignment for the \oursettingtwo{} \oursettingone{} setting still remains an open question.

While highly effective for single-turn instruction following, extending RLHF to multi-turn dialogue introduces significant challenges. These include capturing context-dependent preferences that evolve over the conversation, maintaining long-term coherence and consistency, the increased cost and complexity of collecting high-quality multi-turn preference data, and potential reward hacking where the model optimizes for local turn-level rewards at the expense of overall conversational quality \cite{bai2022training}.

Some works take an initial step toward multi-turn alignment by leveraging conversation-level human feedback in purely textual multi-turn dialogue, mainly focusing on \textit{how to generate better multi-turn dialogue} \citep{shani2024multi}. However, improving instruction-following abilities for \textit{multi-modal understanding and generation} in \textit{multi-turn} settings still remains an open challenge.

\subsection{Evaluating Multi-turn Multimodal Capabilities}

Recent advancements in evaluating multi-turn multimodal capabilities of MLLMs have highlighted the need for benchmarks that reflect real-world conversational complexities. Traditional evaluation datasets often focus on single-turn interactions or unimodal inputs, which do not adequately capture the challenges posed by multi-turn, multimodal dialogues.

To address this gap, several benchmarks have been proposed \citep{xu2024lvlm,liu2024mmdu,epstein2024mmmt,li2024seed,fu2023mme,huang2024dialoggen,hahn2025proactive,zhou2024gate}. For instance, MMDU introduces a comprehensive benchmark designed to evaluate MLLMs' abilities in multi-turn and multi-image conversations  \citep{liu2024mmdu}. It emphasizes the importance of long-context understanding and the integration of multiple images within a single dialogue, pushing models to handle more realistic and complex interactions. 
ConvBench introduces a hierarchical evaluation framework that assesses LVLMs across three cognitive levels: perception, reasoning, and creativity \citep{liu2024convbench}. This structure enables a nuanced analysis of model performance in multi-turn dialogues, highlighting specific areas for improvement.
Similarly, MMMT-IF presents a challenging benchmark focusing on instruction-following in multimodal, multi-turn dialogues  \citep{epstein2024mmmt}. It introduces the Programmatic Instruction Following (PIF) metric, which assesses a model's ability to follow instructions dispersed across long dialogues, requiring the retrieval and reasoning over instructions spread throughout the context. 
In the realm of language models, MT-Eval offers a comprehensive benchmark to evaluate multi-turn conversational abilities \citep{kwan2024mt}. By analyzing human-LLM conversations, it categorizes interaction patterns and constructs multi-turn queries to assess models' performance in maintaining context and coherence over multiple turns.
Collectively, these benchmarks highlight the importance of developing evaluation methods that reflect the intricacies of multi-turn, multimodal interactions. However, a common limitation among them is the primary focus on understanding capabilities, often neglecting the generation aspect of multimodal interleaved information. This oversight presents challenges in providing per-turn and overall feedback judgments, which are crucial for the comprehensive assessment and improvement of MLLMs.

\section{Data Examples}
\label{app:data_examples}

We conduct an in-depth comparison of both open-source and API-based models, including Janus~\citep{wu2024janus,chen2025janus,ma2024janusflow} and Gemini~\citep{reid2024gemini,team2023gemini}, on multi-turn multimodal understanding and generation tasks (Case Study). We further present representative examples of multi-turn QA and preference-annotated instances in \ours{} (Examples). Please refer to \url{https://pku-intermt.github.io/} for more details.

\section{Data Details}
\label{app:data_details}

\subsection{Existing Asset Licenses}

The \ours{} dataset is released under the \textbf{CC BY-NC 4.0} License.
Some seed questions used for eliciting multi-turn communications are sourced from open-source datasets, as shown in Table \ref{tab:app_subset_datasets}, all of which are also under the \textbf{CC BY-NC 4.0} License. Additionally, we have obtained data from \href{https://www.wikihow.com}{Wikihow} and \href{https://www.ehow.com}{Ehow} through legitimate means. The real images included in our dataset are sourced from \href{https://www.google.com/imghp}{Google Images} and \href{https://www.pinterest.com/}{Pinterest}, all of which were acquired legally.

\subsection{Data Access}

Our homepage is available at \url{https://pku-intermt.github.io/}. The dataset consists of three parts hosted on Huggingface:

\begin{itemize}[left=0.3em]
\item \ours{}: A human preference dataset contains 15,604 unique seed questions across diverse categories, 52.6k multi-turn interleaved vision-language QA instances, and 32,459 sets of multi-dimensional human preference annotations. It is available at \url{https://huggingface.co/datasets/PKU-Alignment/InterMT}.
\item \oureval{}: A carefully constructed dataset for evaluating MLLMs in assisting judgment capabilities under multi-turn multimodal understanding and generation. It is available at \url{https://github.com/cby-pku/InterMT}.
\item \textsc{InterMT-Judge}: A tool that leverages \ours{} preference modeling for multi-turn multimodal judge scenarios, achieving a consistency rate of 75\%, outperforming most advanced API-based models. It is available at \url{https://huggingface.co/PKU-Alignment/InterMT-Judge}.
\end{itemize}

\subsection{Instituional Review Board (IRB)}
The human annotations and data usage in this
work have received approval from the Institutional Review Board (IRB) of the Institute for Artificial
Intelligence at Peking University, and the relevant materials are included in the supplementary files.

\paragraph{Fair and Ethical Labor}
We employed 30 full-time crowdsourced workers with substantial experience in multimodal annotation for leading commercial language models. To acknowledge their contributions, we adopted a fair and transparent compensation scheme. The estimated average hourly wage ranged from USD 8.56 to USD 10.23 (XE rate as of 2025/05/13), substantially exceeding the local minimum wage of USD 3.66 in Beijing, PRC \citep{noauthor_undated-nz}. In accordance with local labor laws, workers followed a standard Monday-to-Friday schedule, working eight hours per day with weekends off.

\paragraph{Fair Use of Dataset and Identifying Potential Negative Societal Impacts}
The \ours{} project has undergone a thorough review and audit by the Academic Committee of the Institution for Artificial Intelligence at Peking University. An Institutional Review Board (IRB) has evaluated this work to ensure that the use of the \ours{} dataset adheres to principles of fairness and integrity. 
During dataset construction, we conducted NSFW filtering to enhance internal safety; however, we acknowledge that absolute safety cannot be guaranteed. Given that multimodal data may pose greater societal risks than pure text data, we believe it is necessary to consider implementing safeguards for sensitive content, such as adopting Hugging Face’s gated dataset access settings.
We are committed to developing safe and beneficial AI technologies and strongly oppose any misuse that hinders human progress. We unequivocally condemn malicious use of the \ours{} dataset and advocate for its responsible and ethical use.

\subsection{Comparison with other datasets}

As shown in Table \ref{tab:dataset_comparison}, compared to existing multimodal datasets, \ours{} is the first human preference dataset designed for multi-turn multimodal interactions. Each multi-turn QA instance includes interleaved textual and visual content in both inputs and outputs, with an average of 5.33 images per conversation, simulating complex real-world human-AI communication scenarios.

\begin{table}[ht]
\centering
\caption{Comparison between \ours{} with other image-text datasets. Inter-I: interleaved image-text input; Inter-O: interleaved output; Multi-I: multi-turn for input; Multi-O: multi-turn for output.}
\resizebox{1.0\textwidth}{!}{
\begin{tabular}{l|ccccccc}
\toprule
\textbf{Dataset} & \textbf{Data Scale} & \textbf{Inter-I} & \textbf{Inter-O} & \textbf{Multi-I} & \textbf{Multi-O} & \textbf{\#Num Categories} & \textbf{Preference}    \\
\midrule
CoMM \citep{chen2024comm} & 227k & Yes & Yes & No & No & 5 & No \\
OBELITICS \citep{laurencon2023obelics} & 141M & Yes & Yes & No & No & 200 & No \\
MMC4 \citep{zhu2023multimodal} & 101.2M  & Yes & Yes & No & No & 30 & No \\
Visual Dialogue \citep{das2017visual} & 120k & Yes & No & Yes & Yes & 80 & No \\
IGC \citep{mostafazadeh2017image} & 4.2k & Yes & No & Yes & Yes & N/A  & No \\
Image-Chat \citep{shuster2018image} &  202k & Yes & No & Yes & Yes & 215 & No \\
MM-Dialogue \citep{feng2022mmdialog} & 1.08M & Yes & Yes & Yes & Yes &  4184 & No \\
RLHF-V \citep{yu2024rlhf} & 5.7k & Yes & No & No & No & - & Yes \\
\midrule
\ours{} (\textbf{Ours}) & 32.4k & Yes & Yes & Yes & Yes & 15+ & Yes  \\
\bottomrule
\end{tabular}
}

\label{tab:dataset_comparison}
\end{table}

\section{Data Collection}
\label{app:data_collection}

In this section, we detail the data construction process of \ours{}, as illustrated in Figure~\ref{fig:multi_turn_qa_construction_framework}. The pipeline consists of four main stages.
\textbf{Stage I:} Seed questions are collected from open-source corpora, web content, and human-authored sources. These are then filtered based on perceived quality, topical diversity, and potential for multi-turn expansion.
\textbf{Stage II:} We apply iterative prompting of MLLMs, augmented with external tools (\textit{e.g.}, editing, retrieval, and generation), to produce answer elaborations and follow-up questions, constructing candidate QA trees.
\textbf{Stage III:} Human annotators perform both per-turn (local) and conversation-level (global) assessments—evaluating dimensions such as quality, coherence, context awareness, and completeness—to prune and select preferred branches.
\textbf{Stage IV:} The selected branches are reorganized into deep, coherent QA trees (with depth $\geq$ 5), forming the final multi-turn QA pairs used for model training.

\subsection{Prompt Collection}
\label{app:data_collection_prompt_generation}

\ours{} is built from 100k image-text QA instances sourced from three primary channels: approximately 72.1\% are derived from open-source corpora—namely, publicly available datasets related to vision-language tasks \citep{ji2024align,zhang2025mm,chen2024comm} (Table~\ref{tab:app_subset_datasets} summarizes the open-corpus vision-language datasets used in our pipeline, along with the input-output formats of their original annotations.) ; around 22.8\% originate from legally scraped web content (\textit{e.g.}, multimodal platforms such as \href{www.wikihow.com}{WikiHow} and \href{https://www.pinterest.com/}{Pinterest}); and the remaining 5.1\% are contributed by researcher-curated, human-written prompts. These instances span a wide range of vision-language tasks, including activity generation, data visualization, and table analysis.

\begin{table}[ht]
\centering
\caption{Collected datasets and their corresponding task types. We select various datasets to ensure that the \textit{seed question}s encompass diverse query styles and originate from a broad range of sources.}
\small
\renewcommand{\arraystretch}{1.2}
\begin{tabular}{@{}>{\centering\arraybackslash}m{0.12\linewidth}
                >{\centering\arraybackslash}m{0.27\linewidth}
                >{\centering\arraybackslash}m{0.27\linewidth}
                >{\centering\arraybackslash}m{0.27\linewidth}@{}}
\toprule
\textbf{I/O Format } & \textbf{TI2T} & \textbf{TI2TI} & \textbf{T2I} \\ \midrule
\textbf{Datasets} & 
\begin{tabular}[t]{@{}l@{}}
LLaVA-Instruct-150K \citep{liu2024improved} \\
ART500K \citep{mao2017deepart} \\
MovieNet \citep{huang2020movienet} \\
RLHF-V \citep{yu2024rlhf} \\
ShareGPT4V \citep{chen2023sharegpt4v}
\end{tabular}
&
\begin{tabular}[t]{@{}l@{}}
LLaVA-Instruct-150K \citep{liu2024improved} \\
RLHF-V \citep{yu2024rlhf} \\
MM-RLHF \citep{zhang2025mm} \\
Align-Anything-200K \citep{ji2024align} \\ 
CoMM \citep{chen2024comm} 
\end{tabular}
&
\begin{tabular}[t]{@{}l@{}}
DiffusionDB \citep{wang2022diffusiondb} \\
MS COCO \citep{chen2015microsoft} \\
HPDv2 \citep{wu2023human} \\
Pick-a-Pic-v2 \citep{kirstain2023pick} \\
Align-Anything-200K \citep{ji2024align}
\end{tabular}
\\
\bottomrule
\end{tabular}
\label{tab:app_subset_datasets}
\end{table}

Grounded in theoretical frameworks from linguistics, human-computer interaction, and cognitive psychology \citep{grice1975logic,grosz1995centering,clark1991grounding,parsing2009speech,traum1995computational}, we identify five prototypical scenarios that commonly lead to multi-turn conversations in real-world multimodal contexts: (1) incomplete or unclear user cognition; (2) follow-up queries prompted by unsatisfactory initial responses; (3) complex tasks that require incremental, stepwise reasoning; (4) open-ended or companion-like dialogic interactions; and (5) cross-modal mismatches arising from latent inconsistencies between image and text modalities or the need for integrated cross-modal reasoning. Table \ref{tab:multi_turn_task_types} presents formal definitions of these sceanarios.

\begin{table}[t]
\centering
\caption{Prototypical scenarios that commonly lead to multi-turn conversations in real-world multimodal contexts, grounded in theories of information retrieval and communication.}
\small
\begin{tabular}{>{\raggedright\arraybackslash}p{3.5cm} p{9.5cm}}
\toprule
\textbf{Task Type} & \textbf{Concept} \\
\midrule
\textbf{Unclear Cognition} & Based on Belkin’s ASK model, users are in an "anomalous state of knowledge" during retrieval \citep{belkin1982ask}. They recognize knowledge gaps but cannot clearly articulate their needs. Multi-turn dialogue assists in clarifying their goals through guided interaction. \\
\hline
\textbf{Repeated Attempts Due to Unsatisfactory Answers} & According to Borlund’s interactive IR model, retrieval is a dynamic, iterative process \citep{borlund2003concept}. Users may re-query after unsatisfactory results. Multi-turn dialogue enables iterative feedback and refinement of information needs. \\
\hline
\textbf{Complex Tasks Requiring Stepwise Progression} & Drawing from Ellis’s behavioral model and Kuhlthau’s ISP model, complex tasks require phased progress \citep{ellis1989behavioral,kuhlthau2004seeking}. Multi-turn dialogue supports task decomposition and information integration, helping users build knowledge step by step. \\
\hline
\textbf{Exploratory or Companion-like Interaction} & Bates’s Berrypicking model illustrates non-linear, evolving information behavior \citep{bates1989design}. Users follow shifting interests rather than fixed goals. Multi-turn dialogue provides contextual guidance and emotional engagement in open-ended exploration. \\
\hline
\textbf{Cross-modal Multiturn Interaction} & This involves integrating language and visual modalities. User needs may be embedded across modalities, requiring multi-turn dialogue to semantically align and interpret multimodal information for accurate understanding and task resolution. \\
\bottomrule
\end{tabular}
\label{tab:multi_turn_task_types}
\end{table}

\begin{figure}[t]
    \includegraphics[width=\columnwidth]{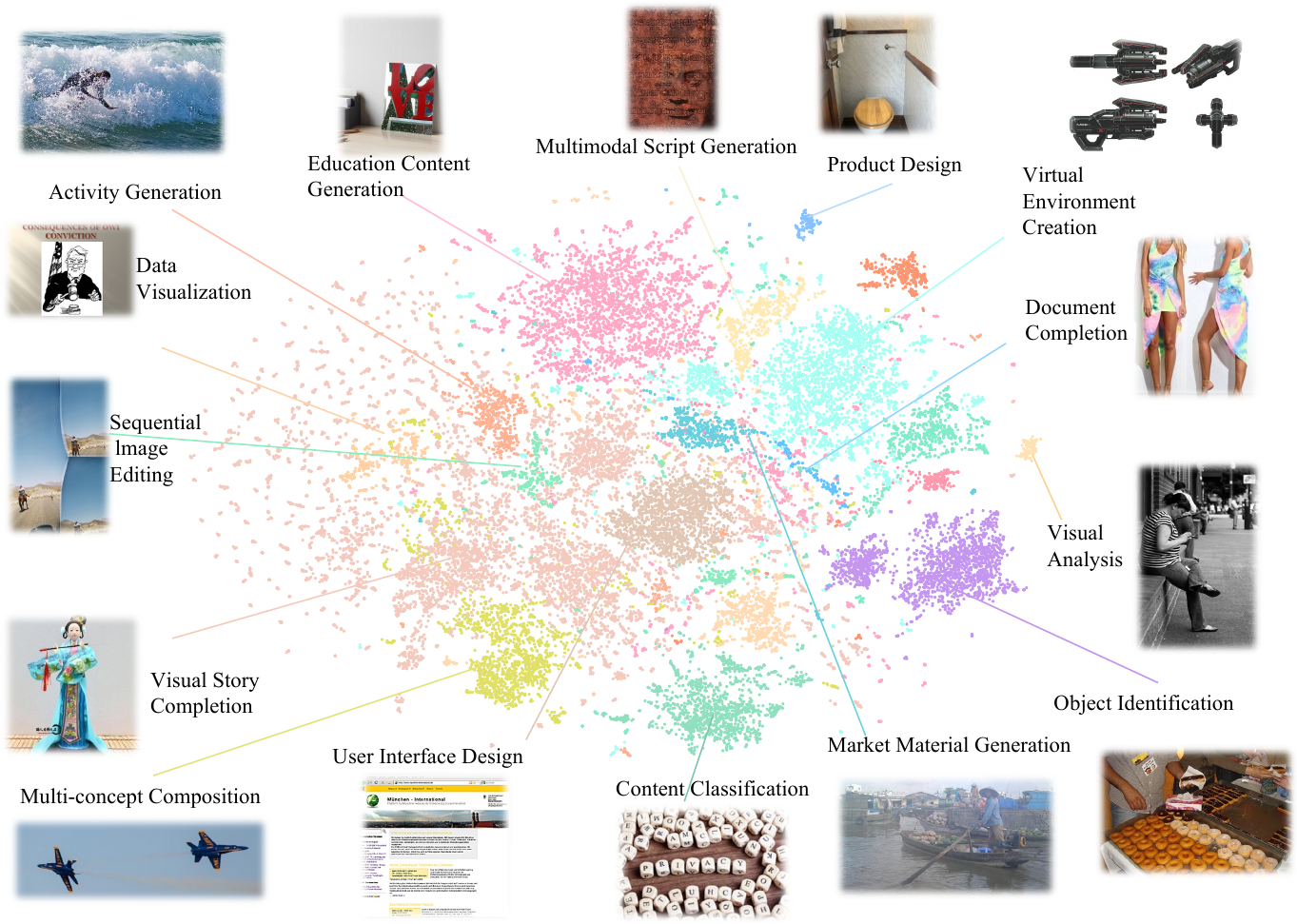}
    \caption{Seed question distribution of \ours{}. Each seed question facilitates \multiturn{multi-turn} QA, encompassing a broad range of over 15 vision-language \understanding{understanding} and \generation{generation} tasks. }
    \label{fig:seed_prompt_distribution}
\end{figure}

Guided by these scenarios, we filter, diversify, and rewrite the original image-text QA instances, resulting in 15,604 unique \textit{seed questions}, which serve as initial prompts for generating iterative, multi-turn conversations. Figure~\ref{fig:multi-turn-dialogue-prompt} presents the system prompt used with GPT-4o \citep{openai2024gpt4o} to evaluate the suitability and potential for multi-turn communication, as well as to assist in filtering and rewriting the original data. Figure~\ref{fig:seed_prompt_distribution} illustrates the distribution of \textit{seed questions} across more than 15 distinct vision-language tasks. Table~\ref{tab:prompt-category} provides definitions and representative examples for each task category.

\begin{table}[ht]
\centering
\caption{Prompt categories and their definitions in multi-turn interleaved multimodal understanding and generation tasks.}
\begin{tabular}{>{\raggedright\arraybackslash}p{3.5cm} p{9.5cm}}
\toprule
\textbf{Prompt Category} & \textbf{Definition} \\
\midrule
\textbf{Fairstyle Generation} & Generate hairstyle designs and styling suggestions based on textual descriptions or reference images. Useful in virtual try-on systems or beauty applications. \\
\hline
\textbf{Report Generation} & Produce structured reports or analytical summaries from multimodal inputs, including images and text. Often used in medical imaging, quality inspection, or news summarization. \\
\hline
\textbf{Activity Generation} & Create interactive activities, games, or engagement schemes tailored to specific topics, scenarios, or user profiles. \\
\hline
\textbf{Document Completion} & Extend or complete existing documents by inferring and preserving their content structure, semantics, and writing style. \\
\hline
\textbf{Visual Story Completion} & Generate coherent continuations or endings for visual narratives based on initial scenes or images. \\
\hline
\textbf{Multimodal Script Generation} & Create instructional or narrative scripts that combine visual and textual components. Common in tutorial videos or AR/VR guides. \\
\hline
\textbf{Sequential Image Editing} & Apply a series of image editing steps in a temporally or logically consistent manner. Suitable for demonstrations or step-wise transformations. \\
\hline
\textbf{Multi-concept Composition} & Integrate multiple concepts, styles, or thematic elements into a unified visual or multimodal output. \\
\hline
\textbf{Education Content Generation} & Generate learning materials, lesson plans, or courseware using multimodal prompts. Can be customized by subject, age group, or learning objectives. \\
\hline
\textbf{Market Material Generation} & Create marketing content such as advertisements, banners, or product showcases leveraging both visual and textual cues. \\
\hline
\textbf{Content Classification} & Organize or categorize multimodal content based on semantic, stylistic, or functional criteria. \\
\hline
\textbf{Visual Analysis} & Analyze visual elements and their interrelations within an image, including object detection, spatial layout, or stylistic attributes. \\
\hline
\textbf{Creative Writing} & Generate creative texts—such as stories, poems, or dialogues—conditioned on visual inputs or scenarios. \\
\hline
\textbf{Technical Explanation} & Provide detailed explanations of technical systems or processes by leveraging both images and text. Often applied in educational or industrial settings. \\
\hline
\textbf{Product Design} & Design new products or optimize existing ones, incorporating visual aesthetics, functionality, and user feedback. \\
\hline
\textbf{Data Visualization} & Translate structured data into visual forms such as charts, diagrams, or infographics to facilitate interpretation. \\
\hline
\textbf{User Interface Design} & Create layouts and elements for user interfaces of digital applications, considering usability and visual coherence. \\
\hline
\textbf{Virtual Environment Creation} & Design and describe immersive virtual spaces or scenes, used in simulation, gaming, or training environments. \\
\hline
\textbf{Other} & User-defined categories not covered above. Allows for flexible extensions based on specific task definitions. \\
\bottomrule
\end{tabular}
\label{tab:prompt-category}
\end{table}

\begin{figure}[htbp]
\begin{minipage}{\textwidth}
    \begin{tcolorbox}[width=\textwidth]
    
    \textbf{System Prompt:}

    You are a communication analysis agent. Your task is to determine whether a given prompt is likely to trigger a \textbf{multi-turn conversation}. Your judgment should be grounded in established discourse and cognitive theories, including Grice’s Cooperative Principle, Centering Theory, Cognitive Load Theory, and known issues in multimodal vision-language alignment.

    Please answer the following questions for each input prompt:

    \textbf{1. Suitability Judgment:} Does the prompt contain characteristics that are likely to elicit follow-up questions, clarification requests, elaboration, or continued user engagement (e.g., due to ambiguity, complexity, or referential uncertainty)? 
    Output either \texttt{YES} or \texttt{NO}.

    \textbf{2. Rationale (if YES):} Briefly explain why this prompt would lead to multi-turn interaction. Your explanation should be based on one or more of the following:

    \begin{itemize}
        \item \textbf{Underspecification or Ambiguity:} The prompt lacks sufficient detail or contains vague references, prompting clarification.
        \item \textbf{Cognitive Complexity:} The task is complex enough to require stepwise reasoning or decomposed planning, encouraging follow-ups.
        \item \textbf{Discourse Dynamics:} Topic or referential focus shifts during the interaction, necessitating communication continuity mechanisms.
        \item \textbf{Multimodal Mismatch:} The prompt involves visual and textual inputs whose alignment must be verified interactively.
        \item \textbf{Exploratory Intent:} The prompt expresses a subjective or open-ended goal, inviting elaboration, negotiation, or perspective sharing.
    \end{itemize}

    \textbf{Provide your answer in the following format:} 

    \texttt{Judgment: [YES/NO]} 

    \texttt{Rationale: [Your explanation here]} 

    \vspace{1em}

    \textbf{User Prompt:}

    Prompt: \{\ldots\}, Image: <image>, your evaluation:
    
    \end{tcolorbox}
\end{minipage}

\caption{The prompt for evaluating the suitability and potential of multi-turn communication. This prompt assesses whether an input is likely to elicit multi-turn interactions and provides theoretical justifications grounded in discourse, cognitive, and multimodal communication theories.}
\label{fig:multi-turn-dialogue-prompt}

\end{figure}

\subsection{Iterative Questioning and Response Generation}

\paragraph{Iterative Questioning} 
To simulate realistic multi-turn communications, the construction process begins with carefully designed \textit{seed questions} that possess the potential to trigger extended multimodal conversations. In subsequent rounds, agents adopt a \textit{Socratic questioning} strategy, generating context-aware follow-up questions based on the prior conversation history. These follow-ups fall into five common categories frequently observed in real-world multimodal conversations: emotional responses that convey empathy or affective engagement, inquiries that deepen or elaborate on prior content, challenges that test logical consistency or factual accuracy, task decomposition for complex problem solving, and natural terminations when the topic has been sufficiently explored.
At each turn, a pool of \textit{10} candidate questions is generated by diverse agents, and a subset of $\mathcal{M}$ (typically 1–3) high-quality and low-redundancy candidates is selected based on textual similarity ranking and regular-expression-based filtering of malformed text. The resulting follow-up questions consistently maintain contextual coherence, ensure conversation continuity, and often leverage visual modalities when necessary to enhance clarity or specificity. Figure \ref{fig:followup_question_generation} presents the system and user prompt for generating follow-up questions.

\paragraph{Response Generation} 
In each turn, every follow-up question ($\mathcal{M}$ per round) is addressed by sampling \textit{10+} candidate responses from diverse agent models. Each response is paired with multiple visual candidates, forming a multimodal answer set. Outputs are expected to be complete, accurate, concise, and helpful, with optional user-guided continuations (\textit{e.g.}, \textit{Would you like a further explanation ?}) to improve user satisfaction.
A subset of $\mathcal{N}$ responses (typically 2–4) is selected based on contextual relevance and multimodal quality. Repeating this process across $n$ rounds yields a tree-structured QA dataset, where each seed question expands into $\prod_{i=1}^{n} \mathcal{M}_i \times \mathcal{N}_i$ multi-turn paths. Figure \ref{fig:multimodal_response_generation} presents the system and user prompt for generating multimodal responses.

\begin{figure}[htbp]
\begin{minipage}{\textwidth}
    \begin{tcolorbox}[width=\textwidth]

    \textbf{System Prompt:} 

    You are a large multimodal language model simulating a curious and thoughtful human. You are currently engaged in a conversation with an AI Assistant. You will receive the previous turns of the conversation along with the AI Assistant's latest reply.

    Your task is to ask a follow-up question or respond interactively based on the AI Assistant's most recent response. Before asking a question, you should first try to understand the conversation history and the User's intent to help you generate a better question for the User.

    Then, select one of the following interaction categories that best describes your intent:

    \begin{itemize}
        \item \textbf{[Emotional Response]}: express emotions, empathy, encouragement, or reactive questions  
        \item \textbf{[Follow-up]}: dig deeper or extend the previous answer  
        \item \textbf{[Challenge]}: question the logic, detail, or validity of the answer  
        \item \textbf{[Step-by-step Task]}: break down a complex task and guide to the next step  
        \item \textbf{[END]}: choose to end the conversation when the topic has been fully explored  
        \item \textbf{[Other: XXX]}: define your own category if needed  
    \end{itemize}

    If helpful, you may reference selected modalities to support your question. Use the following format to include them:
    \texttt{<Modality, brief description>}  
    Examples: \texttt{<Image, diagram of a volcano>}, \texttt{<Audio, sound of rain>}, \texttt{<Video, cat jumping over a box>}

    \textbf{Output Format:}

    \texttt{[[Category]][[Your Question]]}

    \textbf{Examples:} 

    \texttt{[[Follow-up]][[You mentioned that volcanic eruptions are often preceded by earthquakes. Can we use seismic data to predict eruptions in advance?]]}

    If you believe no further question is necessary, conclude the conversation with: 
    \texttt{[[END]][[Some words to end the conversation]]}

    \vspace{1em}

    \textbf{User Prompt:}

    \texttt{Chat History: \{chat\_history\}}

    \texttt{Selected Modalities: \{selected\_modalities\} (default = text,image)}

    \texttt{Last Turn Response: \{last\_turn\_response\}}

    \end{tcolorbox}
\end{minipage}

\caption{System and user prompts for follow-up question generation.}
\label{fig:followup_question_generation}

\end{figure}

\begin{figure}[htbp]
\begin{minipage}{\textwidth}
    \begin{tcolorbox}[width=\textwidth]

    \textbf{System Prompt:} 

    You are a multimodal AI assistant. Your job is to generate helpful, engaging, and clear responses based on user input, which may include text, images, audio, or video.

    \textbf{Instructions:}

    \begin{itemize}
        \item Understand the user's intent by analyzing \textbf{all input modalities}.
        \item Provide a \textbf{complete, accurate, concise, and helpful} response.
        \item \textbf{Use multimodal outputs purposefully}, to enhance clarity, immersion, or user experience.
        \item If the \texttt{AllowedModalities} list includes non-text types, incorporate \textbf{at least one} of them when relevant.
        \item Clearly mark non-text content using: \texttt{<[Modality], brief description>}
        \item Examples: \texttt{<Image, diagram of a volcano>}, \texttt{<Audio, sound of rain>}, \texttt{<Video, cat jumping over a box>}
        \item \textbf{Optionally conclude} your response with a natural follow-up question or suggestion to \textbf{encourage multi-turn conversation}.
        \item Besides user question, you will also receive a list of previous user questions and assistant responses (chat history). You should base your response on the chat history.
        \item You should also consider the user's intent and the chat history when generating your response.
    \end{itemize}

    \textbf{Modality Control:}

    \begin{itemize}
        \item Only use modalities listed in \texttt{AllowedModalities}.
        \item If \texttt{AllowedModalities = []}, generate a text-only response and briefly explain why no other modality is included.
        \item Never fabricate modality content or reference unsupported types.
    \end{itemize}

    \textbf{Input may include:}
    \begin{itemize}
        \item A text prompt
        \item Optional and random input modalities for the user prompt (image, audio, video)
    \end{itemize}

    Always ground your response in the actual input provided.

    \vspace{1em}

    \textbf{User Prompt:}

    \texttt{User Prompt: \{prompt\}}

    \texttt{Previous User Questions and Assistant Responses: \{chat\_history\}}

    \texttt{AllowedModalities: \{allowed\_modalities\} (default = text,image)}

    \end{tcolorbox}
\end{minipage}

\caption{System and user prompts for multimodal interactive answer generation.}
\label{fig:multimodal_response_generation}

\end{figure}

\subsection{Agent Construction}

The agent workflow is built upon a combination of strong open-source models (\textit{e.g.}, Qwen2-VL \citep{Qwen2VL}, Qwen2.5-VL \citep{bai2025qwen2}, Gemma3 \citep{team2025gemma}, and LLaVA-1.5 \citep{liu2024llava} series) alongside leading API-based models (\textit{e.g.}, GPT-4o \citep{openai2024gpt4o}, Gemini-2.0-Flash \citep{google2025gemini2flash} and Claude-3.7-Sonnet-Thinking \citep{anthropic2024claude3}).

Specifically, the following model list are used to construct agents for iterative question generation and response: API-based models include GPT-4o \citep{openai2024gpt4o}, Gemini 2.0 Flash \citep{google2025gemini2flash}, Claude 3.7 Sonnet (both thinking and standard variants) \citep{anthropic2024claude3}. Open-source models include Qwen-2-VL-72B-Instruct \citep{Qwen2VL},  Qwen2.5-VL-32B-Instruct \citep{bai2025qwen2}, Gemma3-27B-Instruct \citep{team2025gemma}, and LLaVA-v1.5-7B \citep{liu2023visual}.

To support diverse multimodal operations, three types of image-centric tools are integrated: (1) text-to-image generators (\textit{e.g.}, FLUX.1-Schnell \citep{black2024flux} and Stable-Diffusion \citep{rombach2021highresolution}) for producing high-quality images based on prompts; (2) an image editing API (\textit{e.g.}, Gemini-2.0-flash-exp-image-generation \citep{google2025geminiimagegeneration}) capable of cropping, highlighting, and modifying images; and (3) web-based retrieval interfaces (\textit{e.g.}, \href{https://images.google.com/}{Google Images}, \href{https://pinterest.com/}{Pinterest}) for sourcing real-world visuals.

\subsection{Quality Control and Pruning}
We employ a multi-perspective filtering strategy to ensure the quality and coherence of the dataset, which can be broadly categorized into two types.

\begin{itemize}[left=0.3em]
    \item \textbf{Image(-Text) Filter}: For single-turn image selection, both visual quality and semantic consistency with the text are critical to ensure the selected image is both legible and contextually appropriate. We adopt an image(-text) filter that integrates visual quality assessment and semantic alignment with the input text to rank and filter the candidate image pool returned by the image tool calling module. Specifically, given candidate images $I = \{i_1, \dots, i_N\}$ and caption $T$, we assign each image a score that combines two factors: a rule-based score $\text{Rule}_j$ combining multi-dimensions (\textit{e.g.}, resolution, clarity \textit{etc.}), and a semantic coherence score $\widetilde{\text{Coher}}_j$ measuring CLIP-based image--text similarity~\citep{radford2021learning}. The final score is computed as:
    \begin{equation}
    S(i_j, T) = \alpha\,\text{Rule}_j + (1 - \alpha)\,\widetilde{\text{Coher}}_j.
    \end{equation}
    Finally, the image $i_{j^*}$ with the highest score is selected, striking a balance between visual quality and semantic coherence to ensure the image is both visually appealing and contextually appropriate.
    \item \textbf{Consistency Filter}: In multi-turn conversations, consistency with prior turns is crucial, both in content (avoiding contradictions with chat history) and style (\textit{e.g.}, maintaining uniform image aesthetics across turns). Advanced models (\textit{e.g.}, GPT-4o~\citep{openai2024gpt4o} and Gemini-2.0-Flash~\citep{google2025gemini2flash}) are employed to better capture such dependencies and ensure coherent filtering across turns.
\end{itemize}
We then prune the generated tree-structured multi-turn paths based on overall image quality, sequence coherence, and diversity. Paths that include irrelevant images or excessively divergent follow-up questions are removed, resulting in a refined set of multi-turn QA instances for human annotation.

\paragraph{Rule-based Filtering}
Given a set of candidate images $I=\{i_1,\dots,i_N\}$ and an associated text description $T$, we first extract for each successfully loaded image $i_j$ a collection of raw quality metrics:
\[
  x_j \,\in\,
  \bigl\{Res_j,\,Clar_j,\,Bright_j,\,Cont_j,\,Color_j\bigr\},
\]
where
\begin{align}
  Res_j    &= W_j \times H_j,\\
  Clar_j   &= \mathrm{Var}\bigl(\mathrm{Laplacian}(\mathrm{Gray}(D_j))\bigr),\\
  Bright_j &= \mathrm{Mean}\bigl(\mathrm{Gray}(D_j)\bigr),\\
  Cont_j   &= \mathrm{Std}\bigl(\mathrm{Gray}(D_j)\bigr),\\
  Color_j  &= \sqrt{\bigl(\mathrm{Std}(R_j - G_j)\bigr)^2
             + \bigl(\mathrm{Std}(0.5(R_j+G_j)-B_j)\bigr)^2}
  \nonumber\\[-2pt]
           &\quad\;+\;0.3\,\sqrt{\bigl(\mathrm{Mean}(R_j - G_j)\bigr)^2
             + \bigl(\mathrm{Mean}(0.5(R_j+G_j)-B_j)\bigr)^2}\,.
\end{align}

\paragraph{Min–Max Normalization.}  
Each metric is normalized to $[0,1]$ via
\[
  \widetilde{x}_j
  = \frac{x_j - \min_{i} x_i}{\max_{i} x_i - \min_{i} x_i}.
\]
Denote the normalized scores $\widetilde{Res}_j,\widetilde{Clar}_j,\widetilde{Bright}_j,\widetilde{Cont}_j,\widetilde{Color}_j$.

\paragraph{Brightness Penalty.}  
We impose a smooth penalty proportional to deviation from the acceptable brightness range $[30,220]$:
\[
  Pen_j
  = -200 \cdot \Bigl[\max(0,30 - Bright_j) + \max(0,Bright_j - 220)\Bigr],
\]
and then normalize
\[
  \widetilde{Pen}_j
  = \frac{Pen_j - \min_i Pen_i}{\max_i Pen_i - \min_i Pen_i}
  \quad\bigl(\widetilde{Pen}_j\in[0,1]\bigr).
\]

\paragraph{Rule-Based Quality Score.}  
The normalized quality score is a weighted sum of the normalized metrics plus the penalty.
We set fixed weights summing to 1:
\[
  w_1 = 0.20,\quad
  w_2 = 0.25,\quad
  w_3 = 0.15,\quad
  w_4 = 0.25,\quad
  w_5 = 0.15,
  \quad
  \sum_{k=1}^5 w_k = 1.
\]
Thus the normalized quality score is
\begin{equation}
  Rule_j
  = 0.20\,\widetilde{Res}_j
  +0.25\,\widetilde{Clar}_j
  +0.15\,\widetilde{Cont}_j
  +0.25\,\widetilde{Color}_j
  +0.15\,\widetilde{Pen}_j.
\end{equation}

\paragraph{Text–Image Coherence.}  
We use ~\href{https://huggingface.co/openai/clip-vit-base-patch32}{\texttt{openai/clip-vit-base-patch32}} to compute a CLIP-based similarity
\[
  Coher_j = \mathrm{CLIP\_score}(D_j, T),
  \quad
  Coher_j\in[-1,1]\;
\]

Then the raw CLIP similarity $Coher_j\in[-1,1]$ is shifted and scaled to $[0,1]$ by
\[
  \widetilde{Coher}_j = \frac{Coher_j + 1}{2}.
\]

\paragraph{Final Selection.} 
Combining normalized quality and coherence, the total score is
\begin{equation}
  S(i_j,T)
  = \alpha\,Rule_j + (1-\alpha)\,\widetilde{Coher}_j,
  \quad
  \alpha\in[0,1].
\end{equation}

We fix the balance parameter $\alpha=0.7$ (putting 70\% weight on visual quality and 30\% on semantic coherence). Finally, we select
\[
  j^* = \arg\max\nolimits_j S(i_j,T),
  \quad
  i_{j^*}\text{ as the output (fallback to }i_1\text{ if all loads fail).}
\]

Thus, each $i_j$ is evaluated both on intrinsic visual quality and on semantic alignment with $T$, and the maximum‐scoring image is selected.

\subsection{Human Preference Annotation}

Defining high-quality multi-turn multimodal communications is inherently challenging, as it involves evaluating response accuracy, the coherence of image-text interactions, and the evolving nature of human preferences over the course of the conversation. We conduct multiple rounds of in-depth discussions with our annotation team, regarding existing open-source datasets and prior work on MLLMs.
We then identify three key criteria: (1) \textit{image-text coherence and helpfulness} — responses should align well with visual content and be logically complete; (2) \textit{contextual consistency} — each turn should maintain thematic relevance, preserve core topics, and ensure stylistic continuity; (3) \textit{long-horizon evaluation} — both local (turn-level) and global (conversation-level) quality should be assessed to evaluate each turn’s contribution to overall conversation.

\begin{multicols}{2}
\begin{itemize}[left=0cm]
\setlength\itemsep{-0.25em}
 \item G1: Context Awareness
 \item G2: Helpfulenss and Completeness
 \item G3: Crucial Step Recognition
 \item G4: Global Image-Text Consistency
 \item G5: Style Coherence
 \item L1: Local Image-Text Consistency
 \item L2: Visual Perceptual Quality
 \item L3: Contextual Coherence 
 \item L4: Text Quality
\end{itemize}
\end{multicols}

Based on these principles\footnote{$G_i$ is used for global evaluation, and $L_i$ is for local evaluation.}, we evaluate multi-turn QA instances from both local and global perspectives. Crowdworkers first rate single turns across four sub-dimensions, then assess the full conversation across five dimensions, finally providing preference labels based on aggregated scores.

\paragraph{Dual Verification}
All annotations are first completed by a dedicated full-time annotation team and subsequently reviewed by a professional quality control unit, which collaborates closely with our researchers to ensure guideline adherence. Additionally, our team manually audits 20\% of the data. Although the task involves inherently subjective human judgments, this dual verification stage primarily aims to suppress annotation noise and improve data quality. Appendix \ref{app:annotation_documents} presents the annotation documents.

\begin{table}[ht]
\centering
\caption{\textbf{Human agreement across different sub-dimensions.}}
\footnotesize
\resizebox{\textwidth}{!}{
\begin{tabular}{cccccccccc}
\toprule
 & \textbf{G1} & \textbf{G2} & \textbf{G3} & \textbf{G4} & \textbf{G5} & \textbf{L1} & \textbf{L2} & \textbf{L3} & \textbf{L4} \\ \midrule
\textbf{w/o Language Feedback (\%)} & 82.1 ± 2.0 & 81.4 ± 2.8 & 83.7 ± 3.5 & 80.6 ± 2.3 & 83.2 ± 2.5 & 82.8 ± 3.1 & 84.1 ± 2.7 & 85.9 ± 2.6 & 86.3 ± 3.0 \\ \midrule
\textbf{w/ Language Feedback (\%)} & -- & -- & \textbf{88.3 ± 1.3} & -- & -- & \textbf{87.2 ± 1.5} & \textbf{85.8 ± 1.6} & -- & -- \\ \bottomrule
\end{tabular}
}
\label{tab:human_preference_agreement}
\end{table}

\paragraph{One More Thing: Language Feedback}

Building on prior works~\citep{dai2024safesora,ji2024align,ji2024aligner}, we incorporate human-written natural language feedback into the annotation process. Each feedback instance includes: (i) \textit{reason}, explaining the rationale behind the assigned score; (ii) \textit{critique}, identifying strengths and weaknesses of the response based on detailed evaluation criteria; and (iii) \textit{refinement}, offering suggestions for improving the image or textual quality of the response. The structured feedback protocol significantly improves inter-annotator agreement by 5–10 percentage points, as shown in Table \ref{tab:human_preference_agreement}.

\subsection{Analysis of Tool Usage Patterns}

 \revision{During the construction of \ours{}, we employ a variety of image-related tools, including image generation, editing, and retrieval, to enable backbone models to simulate multimodal, multi-turn input-output scenarios across partial tasks.
We analyze the proportion of tool calls across different data categories\footnote{Thanks to Reviewer dyyF for inspiring this analysis.}. Table~\ref{tab:tool_usage} summarizes a subset of our results.}

\begin{table}[h]
\centering
\caption{Tool usage distribution by categories in \ours{}.}
\label{tab:tool_usage}
\begin{tabular}{lccc}
\toprule
Category & Edit & Retrieve & Generate \\
\midrule
Visual Story Completion & 0.12 & 0.29 & 0.59 \\
Multi-concept Composition & 0.23 & 0.31 & 0.46 \\
Content Classification & 0.11 & 0.63 & 0.26 \\
Visual Analysis & 0.34 & 0.47 & 0.19 \\
Sequential Image Editing & 0.72 & 0.18 & 0.10 \\
\bottomrule
\end{tabular}
\end{table}

We observe distinct patterns of tool usage across task categories. For instance, image editing tools dominate sequential image editing tasks, while generation and retrieval tools primarily assist user comprehension. In contrast, all three tools are broadly employed in image tutorial and multi-turn knowledge QA tasks.

\section{Annotation Documents}
\label{app:annotation_documents}
\subsection{Withdraw}
\paragraph{What is an incorrect answer?}
\begin{itemize}
    \item Providing a link that cannot be accessed.
    \item Giving the current date, but it does not match the actual date.
    \item Providing a highly time-sensitive response, while the actual situation has changed. For example, listing the "top ten trending songs" without specifying the inability to access the most recent data will be considered invalid.
    \item Factual errors that contradict objective reality.
\end{itemize}

\paragraph{What is an invalid question?}
We carefully examine invalid questions during data validation and continuously update the definition of invalid questions.
\begin{itemize}
    \item Incomplete questions, such as containing only a single word like "I" or "Hello."
    \item Questions that lack context, making them difficult to understand.
    \item Requests for analysis of a given text or context without actually providing the text or context.
    \item Questions that contain factual errors, rendering the question itself invalid.
\end{itemize}

\paragraph{What is an ungradable question?}
\begin{itemize}
    \item Highly subjective tasks, such as creative writing, where there is no objective standard for determining quality.
    \item Questions that exceed the annotator's knowledge level, such as those involving advanced coding, finance, computer science, or physical laws.
    \item Two questions with answers that are too similar, such as "apple" and "apple." (only differing by punctuation).
\end{itemize}

\paragraph{What are questions that require web search?}
Many questions require searching the internet, especially when objective facts are needed.

\subsection{Features of Annotated Data}
\paragraph{Core Characteristics}
\begin{itemize}
    \item Multi-turn Dialogue: The dataset includes both short dialogues (2-3 turns) and long dialogues (5 or more turns) to accommodate various task requirements.
    \item Image-Text Interaction: The data includes a combination of text input, image input, and mixed text-image input.
    \item Scoring System: The scoring follows a fine-grained scale, where each dialogue turn is independently scored, and the final composite score reflects overall performance.
\end{itemize}

\paragraph{Data Types}
\begin{itemize}
    \item Image-Text Input with Multi-turn Text Output (e.g., step-by-step optimization of a design image).
    \item Text Input with Multi-turn Image Output (e.g., describing an object and generating images from various perspectives).
    \item Image-Text Input with Multi-turn Image-Text Output (e.g., design modification process with visual outputs).
    \item Image Input with Multi-turn Text Output (e.g., providing detailed interpretation or analysis of an uploaded image).
\end{itemize}

\subsection{Annotation Guidelines}
\subsubsection{Overall Response Evaluation}
\paragraph{Context Awareness}
\textbf{Definition}: The model should retain and understand the dialogue history to ensure contextual coherence, rather than treating each turn as an isolated interaction.

\textbf{Examples}:
\begin{itemize}
    \item In \textit{visual storytelling} tasks, the model should maintain consistent characters, settings, and plot lines.
    \item In \textit{design revision} tasks, the model should remember the user's previous requests and avoid repeating suggestions that were previously rejected.
\end{itemize}

\textbf{Scoring Criteria}:
\begin{itemize}
    \item \textbf{0 points}: The model completely ignores the context; responses are irrelevant or contradict the dialogue history.
    \item \textbf{1 point}: The model partially recalls context but exhibits noticeable information loss or inconsistency in roles.
    \item \textbf{2 points}: Context is mostly preserved, with occasional minor inconsistencies.
    \item \textbf{3 points}: Full understanding of context; responses are logically coherent, with no information loss or contradictions.
\end{itemize}

\paragraph{Helpfulness and Completeness}
\textbf{Definition}: Measures how well the model’s textual and visual outputs follow task instructions and provide complete information to fulfill the user’s request. This also includes the logical structure of the response. In multi-turn image-text interactions, the model should accurately follow all instructions and ultimately deliver a complete solution.

\textbf{Example}:
\begin{itemize}
    \item \textit{Task}: Cake design improvement  
    \begin{itemize}
        \item \textit{User}: Please help me improve this design (uploads image)  
        \item \textit{AI}: Suggests adding frosting (but no new image generated) $\rightarrow$ Deduct points
        \item \textit{User}: Please show me the modified 3D rendering  
        \item \textit{AI}: [Generates an image of the cake with frosting] $\rightarrow$ Full score
    \end{itemize}
\end{itemize}

\textbf{Scoring Criteria}:
\begin{itemize}
    \item \textbf{0 points}: The model fails to meet the user’s needs; responses are irrelevant or severely incorrect, making the task unachievable.
    \item \textbf{1 point}: Partially satisfies the user’s request but lacks critical content or contains major errors that hinder task completion.
    \item \textbf{2 points}: Largely completes the task but has minor omissions or inaccuracies that affect the final outcome.
    \item \textbf{3 points}: Fully and accurately fulfills all user requirements; information is comprehensive, logically structured, and free from errors or omissions.
\end{itemize}

\paragraph{Crucial Step Recognition}
\textbf{Definition}: In multi-turn interactions, the model must accurately identify and complete crucial steps, avoiding irrelevant or incorrect information.

\textbf{Example}:
\begin{itemize}
    \item \textit{Task}: Step-by-step guidance for drawing a cat  
    \begin{itemize}
        \item \textit{Crucial steps}: Sketch outline $\rightarrow$ Refine facial features $\rightarrow$ Adjust proportions $\rightarrow$ Apply color
        \item \textit{Incorrect}: Model asks the user to color before the outline is drawn
        \item \textit{Correct}: Model guides the user through steps in a logical order
    \end{itemize}
\end{itemize}

\textbf{Scoring Criteria}:
\begin{itemize}
    \item \textbf{0 points}: Key steps are entirely incorrect or omitted, preventing task completion.
    \item \textbf{1 point}: Some steps are inaccurate, though the task may still proceed with effort.
    \item \textbf{2 points}: Overall step sequence is reasonable, with minor deviations or logical flaws.
    \item \textbf{3 points}: All crucial steps are correctly identified and ordered, with no redundancy or omissions.
\end{itemize}

\paragraph{Global Image-Text Consistency}
\textbf{Definition}: In multi-turn image-text interactions, textual descriptions should align closely with the generated images. Inconsistencies between text and images, or failing to generate images when required, result in lower scores.

\textbf{Example}:
\begin{itemize}
    \item \textit{Task}: AI-generated interior design plan  
    \begin{itemize}
        \item \textit{User}: Please provide a modern-style living room design  
        \item \textit{AI}: [Generates image, but the style does not match] $\rightarrow$ Deduct points
        \item \textit{User}: Please change the sofa color to dark grey  
        \item \textit{AI}: [Generates image with dark grey sofa] $\rightarrow$ Full score
    \end{itemize}
\end{itemize}

\textbf{Scoring Criteria}:
\begin{itemize}
    \item \textbf{0 points}: Images are completely unrelated to the text, or necessary images are missing.
    \item \textbf{1 point}: Partial relevance, but with significant mismatches (e.g., incorrect color or structure).
    \item \textbf{2 points}: Largely consistent, with minor deviations.
    \item \textbf{3 points}: Perfect alignment between text and images, with no inconsistencies.
\end{itemize}

\paragraph{Style Coherence}
\textbf{Definition}: Assesses the consistency of style and subject representation across generated images, including texture, color harmony, lighting, rendering style, physical properties, clothing, and behavior. It penalizes visual repetition, such as overly similar outputs or duplicated elements within a single image. In multi-turn interactions, generated images should exhibit stylistic coherence across turns, with smooth transitions and no abrupt changes.

\textbf{Special Case}:
\begin{itemize}
    \item If only one turn includes an image while the other does not, visual style coherence is \textbf{not} affected. In such cases, assign a \textbf{default score of 3 points}.
\end{itemize}

\textbf{Scoring Criteria}:
\begin{itemize}
    \item \textbf{–1 point}: The task required image generation, but none was provided.
    \item \textbf{0 points}: Images exhibit entirely different styles, tones, rendering, or subject traits, resulting in visual dissonance.
    \item \textbf{1 point}: Some stylistic or subject consistency, but with clear discrepancies (e.g., sudden tone changes, mismatched rendering, or inconsistent subject traits).
    \item \textbf{2 points}: Style, tone, and subject representation are generally consistent, with minor variations that do not affect overall coherence.
    \item \textbf{3 points}: All images are highly consistent in style, tone, quality, and subject representation; visual transitions are smooth and contextually appropriate.
\end{itemize}

\subsubsection{Turn-level Evaluation Metrics}

\paragraph{Local Image-Text Consistency}
\textbf{Definition}: In a single dialogue turn, the textual description should closely match the generated image(s), ensuring the text accurately reflects the visual content without ambiguity or misleading information.

\textbf{Applicable Scenarios}:
\begin{itemize}
    \item If the turn includes multiple images, evaluate the overall consistency of the text with all images. Individual image feedback can be added as needed (e.g., [3,1] Image 1: accurate; Image 2: inconsistent).
    \item If no image is generated, evaluate based on task requirements:
    \begin{itemize}
        \item If image generation was expected but omitted, assess the inconsistency between the text and the missing visual content.
        \item If the task (e.g., Visual Analysis) does not require image generation, assess consistency between the input image and the text.
    \end{itemize}
    \item Otherwise, default evaluation compares the answer text with the image(s) generated in that turn.
\end{itemize}

\textbf{Scoring Criteria}:
\begin{itemize}
    \item \textbf{0 points}:
    \begin{itemize}
        \item Text is irrelevant to the image(s) or contains major factual errors;
        \item Key descriptions are missing or completely incorrect (e.g., referencing nonexistent objects or scenes);
        \item Text may cause significant misunderstanding.
    \end{itemize}
    \item \textbf{1 point}:
    \begin{itemize}
        \item Text is partially related to the image(s), but includes clear errors or misleading descriptions;
        \item Covers part of the image content but omits or misrepresents key details or relationships;
        \item Reader must infer or adjust understanding to align with the image(s).
    \end{itemize}
    \item \textbf{2 points}:
    \begin{itemize}
        \item Text generally matches the image(s), with minor local inaccuracies (e.g., imprecise attribute descriptions or slight omissions);
        \item Does not hinder overall comprehension, but lacks precision upon close inspection.
    \end{itemize}
    \item \textbf{3 points}:
    \begin{itemize}
        \item Text is highly aligned with the image(s), covering all key elements and details;
        \item Free from factual errors or ambiguity; the description is natural and coherent.
    \end{itemize}
\end{itemize}

\paragraph{Visual Perceptual Quality}
\textbf{Definition}: Evaluates the visual realism, naturalness, and absence of distortion or artifacts in the generated image(s). Focuses on whether the image structure, colors, and composition realistically simulate the physical world, avoiding unnatural artifacts.

\textbf{Applicable Scenarios}:
\begin{itemize}
    \item In multi-image outputs, assign a unified score for overall quality. If image quality varies significantly, provide per-image feedback as needed (e.g., [3,1] Image 1: good; Image 2: distorted).
    \item If no image is generated, assess any image provided in the user prompt. If the image has issues, point them out in the textual answer.
\end{itemize}

\textbf{Scoring Criteria}:
\begin{itemize}
    \item \textbf{0 points}:
    \begin{itemize}
        \item Obvious artifacts (e.g., disconnections, misalignments), severe distortions (e.g., highly unrealistic shapes), or structural errors (e.g., unbalanced proportions, illogical composition);
        \item Unnatural color rendering (e.g., harsh color blocks, abnormal tones);
        \item Lighting does not follow physical laws, severely affecting image recognizability.
    \end{itemize}
    \item \textbf{1 point}:
    \begin{itemize}
        \item Image is mostly recognizable but contains localized severe flaws;
        \item Examples: anatomical errors (e.g., limb dislocation), inconsistent local color (e.g., banding, strong noise), or small rendering failures;
        \item Overall naturalness is compromised, affecting visual coherence.
    \end{itemize}
    \item \textbf{2 points}:
    \begin{itemize}
        \item Image is generally natural and coherent; structure, color, and lighting are mostly reasonable;
        \item Minor local imperfections such as rough edges, small artifacts, or slight blurring that do not affect overall perceptual quality.
    \end{itemize}
    \item \textbf{3 points}:
    \begin{itemize}
        \item Image is visually realistic and natural;
        \item Well-structured, smooth color transitions, physically consistent lighting;
        \item No visible artifacts, distortions, or flaws; overall aesthetics and details are excellent.
    \end{itemize}
\end{itemize}

\paragraph{Text Quality}
\textbf{Definition}: Measures the clarity, coherence, and correctness of the output text. Includes grammar, spelling, readability, consistency with instructions and context, and absence of redundancy. Responses should be logically sound, well-structured, and clearly expressed, avoiding abrupt transitions or repetition.

\textbf{Scoring Criteria}:
\begin{itemize}
    \item \textbf{0 points}: Text is disorganized, lacks logic, and is hard to understand; may contain numerous grammar or spelling errors or repetitive content.
    \item \textbf{1 point}: Some parts are logically clear, but the text includes noticeable jumps, omissions, or contradictions that hurt overall readability; may include frequent language errors or redundant expressions.
    \item \textbf{2 points}: The overall logic is reasonable and the flow mostly smooth, but there are minor incoherences; some sentences may require optimization to improve readability.
    \item \textbf{3 points}: Text is logically rigorous, clearly expressed, well-organized, and naturally structured; no obvious jumps or repetition; grammar and spelling are correct, providing a good reading experience.
\end{itemize}

\paragraph{Contextual Coherence}
\textbf{Definition}: Assesses whether the response in this turn logically continues the dialogue history and remains consistent with prior content, avoiding contradictions.

\textbf{Scoring Criteria}:
\begin{itemize}
    \item \textbf{0 points}: Completely irrelevant or logically inconsistent with previous context.
    \item \textbf{1 point}: Partially relevant but includes clear inconsistencies.
    \item \textbf{2 points}: Mostly coherent, with minor deviations.
    \item \textbf{3 points}: Fully consistent with prior dialogue; no contradictions.
\end{itemize}

\section{More details of Annotation}
\label{app:more_details_of_annotation}

\subsection{Annotation Platform}

\begin{figure}[htbp]
    \includegraphics[width=\columnwidth]{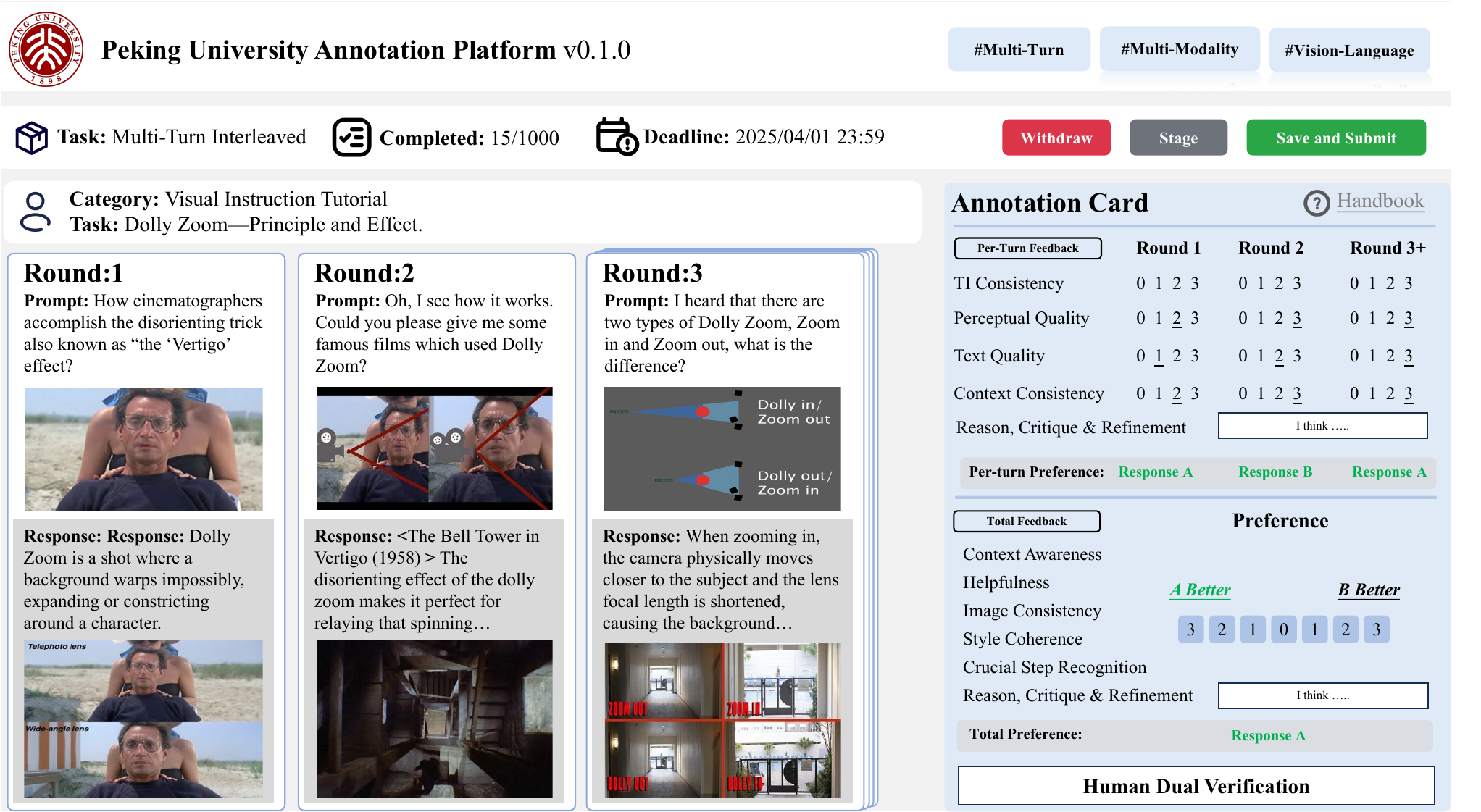}
    \caption{The WebUI of the annotation platform.}
    \label{fig:annotation_platform}
\end{figure}

The annotation platform of \ours{} is similar to our sister projects, PKU-SafeRLHF and BeaverTails.
Based on the specific annotation requirements, we have made appropriate adjustments to the platform, such as adding support for multimodal dialogue inputs across multiple rounds and enabling scoring and preference ranking for each round of communications, as shown in Figure \ref{fig:annotation_platform}. After human annotations, we provide dual verification from both human experts and our researchers.

On the annotation platform, we have provided a comprehensive handbook that includes detailed documentation for the annotation process, as shown in Appendix \ref{app:annotation_documents}, along with summaries and explanations for contentious annotation cases. A withdrawal button is available at the top-right corner of the interface to filter out invalid or meaningless annotation pairs.

Overall, the annotation process consists of three stages: 
\begin{itemize}[left=0.3em]
    \item \textbf{Stage I:} Annotators carefully read and learn the annotation guidelines. They first score each round of a multimodal dialogue (fine-grained scoring) and then determine which round is better between two dialogues (preference ranking).
    \item \textbf{Stage II:} Subsequently, based on the individual round annotations, annotators perform fine-grained scoring for the complete multi-turn multimodal communications and rank which of the two communications is better.
    \item \textbf{Stage III:} The annotation results undergo dual verification by human experts. Annotations with a low consistency rate are rejected for re-annotation. Qualified annotations are then reviewed by researchers through sampling and auditing. Finally, the human preference dataset is finalized.
\end{itemize}

\subsection{Details on Data Labeling Services}
Building upon the successes of previous projects such as BeaverTails \citep{ji2023beavertails}, SafeSora \citep{dai2024safesora}, PKU-SafeRLHF \citep{ji2024pku}, and Aligner \citep{ji2024aligner}, we again collaborated with the professional annotation service provider~\href{www.aijetdata.com}{AIJet Data}. We did not directly interact with the crowdworkers; instead, AIJet managed the entire annotation process. Capitalizing on their expertise in annotating textual data, AIJet curated a dedicated team of experienced annotators tailored to the needs of our project. In light of the task’s complexity, we established a contract with a rate above the industry average to prioritize the engagement of qualified personnel. To ensure consistent annotation quality, we supplied AIJet with comprehensive guidelines aimed at standardizing and refining the labeling criteria.

\subsection{Impact of Image Inputs on Dialogue Dynamics}

\revision{In \ours{}, images simulate realistic user inputs and enhance dialogue understanding and preference annotation in three key ways:}

\begin{itemize}[left=0.1em]
    \item \textbf{Visual Context Improves Queries:} Images prompt richer and more complex questions, reflecting real-world interactions.
    \item \textbf{Task-Specific Relevance:} In domains such as data visualization or medical imaging, visual signals convey information that text alone cannot.
    \item \textbf{Quality Drop Without Images:} Without images, average turns drop from 4.87 to 3.66, and average response length decreases from 874.96 to 402.03 tokens.
\end{itemize}

To quantify this effect, we conduct an ablation study summarized in Table~\ref{tab:image_impact}.

\begin{table}[h]
\centering
\caption{Impact of image inputs on dialogue dynamics.}
\label{tab:image_impact}
\begin{tabular}{lcc}
\toprule
Condition & Average Turns & Average Response Length (tokens) \\
\midrule
With Image Input & 4.87 & 874.96 \\
Without Image Input & 3.66 & 402.03 \\
\bottomrule
\end{tabular}
\end{table}

These results highlight the vital role of images in multimodal preference modeling and validate \ours{}’s design for realistic evaluation.

\subsection{Potential of Extending the InterMT Workflow to Other Modalities}

\revision{While the current implementation of InterMT focuses on image-text interactions, its underlying design principles naturally extend to a broader range of modalities. In particular, the agent-based workflow and preference modeling architecture are modality-agnostic, providing a flexible foundation for incorporating audio, video, and even haptic or sensor-based inputs. This extensibility stems from the framework’s emphasis on multi-turn, fine-grained human preference modeling, which captures interactional dynamics rather than relying on modality-specific representations.}

\revision{Future integration of audio and video modalities would enable the framework to more comprehensively reflect the multimodal nature of human communication, capturing not only semantic content but also prosodic, visual, and affective cues that are critical to real-world understanding and alignment. As large foundation models continue to advance toward unified multimodal reasoning, InterMT workflow can serve as a scalable scaffold for evaluating and aligning these capabilities under complex, multi-agent interaction settings. Ultimately, we envision InterMT as a general-purpose paradigm for studying human-AI alignment across modalities, bridging the gap between symbolic dialogue and embodied communication.}

\subsection{On the Gap Between AI-Synthesized and Human-Human Dialogues}
\label{sec:data_rationale}

\revision{While dialogues authored entirely by humans represent a gold standard, we opt for an AI-synthesis approach after initial experiments reveal significant practical challenges. We identify a core trade-off in human data collection: minimal annotation guidelines often yield brief, underspecified conversations, whereas strict instructions lead to unnatural and homogeneous dialogues as annotators converge on formulaic patterns to meet requirements. Furthermore, the cost is prohibitive; generating high-quality multi-turn dialogues proves to be 7--10$\times$ more time-consuming than standard preference annotation, with costs escalating by over 10$\times$ when including manually sourced or created images.}

\revision{Consequently, we adopt a strategy that leverages AI for generation and humans for high-level critique and preference annotation, a paradigm consistent with prior work~\cite{zheng2023judging, ostyakova2023chatgpt}. This approach allows us to generate a diverse and creative dataset at a feasible cost, a benefit also noted in recent studies~\cite{achiam2023gpt, yu2024rlaif}. To mitigate the inherent risks of model-generated content, such as systematic bias and quality degradation, we implement a multi-faceted quality control process.}

\revision{First, to enhance content diversity and prevent model-specific artifacts, we employ a multi-source generation pipeline. This involves using a mix of closed- and open-source models for language, tool-use, and multimodal understanding, alongside various image generation models (\textit{e.g.}, FLUX, SD-3). We also incorporate a human-in-the-loop diversity audit, where annotators identify and report systematic issues (\textit{e.g.}, repetitive response patterns or image styles), enabling us to refine our models and sampling strategies iteratively.}

\revision{Second, all generated samples undergo a rigorous two-stage validation process. An initial, automated pre-screening stage filters out candidates using rule-based checks on image metrics (\textit{e.g.}, resolution, clarity) and model-based evaluations like CLIP-based image-text similarity~\cite{radford2021learning}. Subsequently, all surviving samples are manually reviewed by our trained annotation team to identify and discard any remaining incoherent or low-quality instances. Only data that successfully pass both automated and human validation are included in the final dataset.}

\section{More Details about \oureval{}}
\label{app:details_of_oureval}
\subsection{Review of Human Annotation Dimensions}

We first revisit the key dimensions of \multiturn{multi-turn} interleaved multimodal \understanding{understanding} and \generation{generation}, which also serve as the annotation criteria for our human-labeled dataset. The evaluation in \oureval{} is conducted with respect to these dimensions, guided by genuine human feedback. Specifically, $G_i$ denotes global evaluation, while $L_i$ corresponds to local evaluation.

\begin{multicols}{2}
\begin{itemize}[left=0cm]
\setlength\itemsep{-0.25em}
 \item G1: Context Awareness
 \item G2: Helpfulenss and Completeness
 \item G3: Crucial Step Recognition
 \item G4: Global Image-Text Consistency
 \item G5: Style Coherence
 \item L1: Local Image-Text Consistency
 \item L2: Visual Perceptual Quality
 \item L3: Contextual Coherence 
 \item L4: Text Quality
\end{itemize}
\end{multicols}

\subsection{Judge Settings and Evaluation Metrics} 

The dataset includes multi-turn multimodal interleaved communication histories and human-annotated ground truth. Evaluated models must assess the conversation at both the turn and conversation levels across nine dimensions, following a set of guidelines. \textit{Scoring Evaluation} requires the model to assign scores on a 0-3 scale, with evaluation based on agreement and Pearson similarity \citep{lee1988thirteen,zheng2023judging,chen2024mllm}. \textit{Pair Comparison} directly compares two individual turns or entire conversations, without considering ties, and is evaluated for accuracy against human judgments. \textit{Crucial Step Recognition} addresses a key challenge in multi-turn conversations: accurately identifying the user’s intent and determining whether it has been fulfilled, evaluated by the score provided by judge according to the human-annotated reference answers.

\paragraph{Note}We use the following notation conventions to refer to proprietary models evaluated in our experiments: \textit{Gemini-Flash*} refers to \texttt{Gemini-2.0-Flash}, \textit{Gemini-Pro*} denotes \texttt{Gemini-2.5-Pro-preview}, and \textit{Claude-thinking*} corresponds to the \texttt{Claude-3.7-Sonnet (thinking)} model.

\subsection{MLLM as a Judge}
\label{app:mllm_as_a_judge}

Inspired by \citep{chen2024mllm,zheng2023judging}, we leverage genuine human-annotated data collected in \ours{} to construct \oureval{}, a benchmark designed to evaluate the alignment between models and human values in multi-turn multimodal interaction scenarios. Our evaluation focuses on three key aspects: \textit{Score Evaluation}, \textit{Pair Comparison}, and \textit{Crucial Step Recognition}. The system and user prompts used for \textit{Score Evaluation} and \textit{Pair Comparison} are illustrated in Figure~\ref{tab:global_judge_prompt_direct}, \ref{tab:local_judge_prompt_direct} and Figure~\ref{tab:global_judge_prompt_direct_preference},\ref{tab:local_judge_direct_preference}, respectively.

We also examine the effect of prompting models to generate rationales on scoring accuracy (Figure \ref{tab:global_judge_prompt_reason}, \ref{tab:local_judge_prompt_reason} and Figure \ref{tab:global_judge_prompt_reason_preference}, \ref{tab:local_judge_reason_preference}. For \textit{Score Evaluation}, we quantify the alignment between model-assigned and human-assigned scores using the Pearson correlation coefficient, computed as follows:

\begin{equation}
r = \frac{\sum_{i=1}^{n} (x_i - \bar{x})(y_i - \bar{y})}{\sqrt{\sum_{i=1}^{n} (x_i - \bar{x})^2} \sqrt{\sum_{i=1}^{n} (y_i - \bar{y})^2}},
\end{equation}

where $x_i$ and $y_i$ denote the scores assigned by the model and human annotators, respectively, and $\bar{x}$ and $\bar{y}$ are their corresponding means.

\begin{figure}[htbp]
\begin{minipage}{\textwidth}
\begin{tcolorbox}[width=\textwidth]

\textbf{System Prompt:}

You are a scoring model for evaluating the overall quality in multi-turn visual dialogues. You will receive a conversation history, please read it carefully. Next, I will provide you with an evaluation list and corresponding scoring criteria. Please score the conversation history based on the scoring criteria. 

Your output needs to be:
\[
[\text{Evaluation Criterion}_1,\ \boxed{\text{score1}}],\ [\text{Evaluation Criterion}_2,\ \boxed{\text{score2}}],\ \ldots
\]

Example:

Evaluation list: 

[context\_awareness,\ helpfulness,\ crucial\_step\_recognition,\ global\_image\_text\_consistency,\ style\_coherence]

Output:

[\text{context\_awareness},\ \boxed{\text{score1}}],\ [\text{helpfulness},\ \boxed{\text{score2}}],\ [\text{crucial\_step\_recognition},\ \boxed{\text{score3}}],\ [\text{global\_image\_text\_consistency},\ \boxed{\text{score4}}],\ [\text{style\_coherence},\ \boxed{\text{score5}}]

\textless \textbf{Annotation Documents}\textgreater

\vspace{1em}
\textbf{User Prompt:}

Now, please evaluate the conversation history based on the scoring criteria. And output the result in the format of:  
\[
[\text{Evaluation Criterion}_1,\ \boxed{\text{score1}}],\ [\text{Evaluation Criterion}_2,\ \boxed{\text{score2}}],\ \ldots
\]

\end{tcolorbox}
\end{minipage}
\caption{System and user prompts for global evaluation in multi-turn multimodal communications (score only).}
\label{tab:global_judge_prompt_direct}
\end{figure}

\begin{figure}[htbp]
\begin{minipage}{\textwidth}
\begin{tcolorbox}[ width=\textwidth]

\textbf{System Prompt:}

You are a scoring model for evaluating the overall quality in multi-turn visual dialogues. You will receive a conversation history, please read it carefully. Next, I will provide you with an evaluation list and corresponding scoring criteria. Please score the conversation history based on the scoring criteria and provide a reason for your score.

Your output needs to be:
\[
[\text{Evaluation Criterion}_1,\ \text{Reason},\ \boxed{\text{score1}}],\ [\text{Evaluation Criterion}_2,\ \text{Reason},\ \boxed{\text{score2}}],\ \ldots
\]

Example:

Evaluation list: 

{[context\_awareness,\ helpfulness,\ crucial\_step\_recognition,\ global\_image\_text\_consistency,\ style\_coherence]}

Output:
\[
[\![\text{context\_awareness},\ \text{Reason},\ \boxed{\text{score1}}],\ [\text{helpfulness},\ \text{Reason},\ \boxed{\text{score2}}],\ \ldots\!]
\]

\textless \textbf{Annotation Documents}\textgreater

\vspace{1em}
\textbf{User Prompt:}

Now, please evaluate the conversation history based on the scoring criteria. And output the result in the format of:  
\[
[\text{Evaluation Criterion}_1,\ \text{Reason},\ \boxed{\text{score1}}],\ [\text{Evaluation Criterion}_2,\ \text{Reason},\ \boxed{\text{score2}}],\ \ldots
\]

\end{tcolorbox}
\end{minipage}
\caption{System and user prompts for global evaluation in multi-turn multimodal communications (with reason).}
\label{tab:global_judge_prompt_reason}
\end{figure}

\begin{figure}[htbp]
\begin{minipage}{\textwidth}
\begin{tcolorbox}[ width=\textwidth]

\textbf{System Prompt:}

You are a scoring model for evaluating the quality of a single turn in multi-turn visual dialogues. You will receive a conversation history, please read it carefully. Next, I will provide you with an evaluation list and corresponding scoring criteria. Please score the conversation history based on the scoring criteria. Your output must follow this exact format:

Evaluation list: 
[\text{local\_image\_text\_consistency},\ \text{visual\_perceptual\_quality},\ \text{text\_quality},\ \text{context\_coherence}]

Output:

[\![\text{local\_image\_text\_consistency},\ \boxed{\text{score}}]\!],\ 
[\![\text{visual\_perceptual\_quality},\ \boxed{\text{score}}]\!],\ 
[\![\text{text\_quality},\ \boxed{\text{score}}]\!],\ 
[\![\text{context\_coherence},\ \boxed{\text{score}}]\!]

Where \texttt{score} is your numerical rating (0--3).

\textless \textbf{Annotation Documents}\textgreater

\vspace{1em}

\textbf{User Prompt:}

Now, please evaluate this turn based on the scoring criteria. Your score should be between 0 and 3. And output the result in the format:
\[
[\![\text{Evaluation Criterion}_1,\ \boxed{\text{score1}}],\ [\text{Evaluation Criterion}_2,\ \boxed{\text{score2}}],\ \ldots\!]
\]

\end{tcolorbox}
\end{minipage}
\caption{System and user prompts for local single-turn evaluation (score only).}
\label{tab:local_judge_prompt_direct}
\end{figure}

\begin{figure}[htbp]
\begin{minipage}{\textwidth}
\begin{tcolorbox}[width=\textwidth]

\textbf{System Prompt:}

You are a scoring model for evaluating the quality of a single turn in multi-turn visual dialogues. You will receive a conversation history, please read it carefully. Next, I will provide you with an evaluation list and corresponding scoring criteria. Please score the conversation history based on the scoring criteria. Your output must follow this exact format:

Evaluation list: 

[\text{local\_image\_text\_consistency},\ \text{visual\_perceptual\_quality},\ \text{text\_quality},\ \text{context\_coherence}]

Output:

[\![\text{local\_image\_text\_consistency},\ \text{reason},\ \boxed{\text{score}}]\!],\ 
[\![\text{visual\_perceptual\_quality},\ \text{reason},\ \boxed{\text{score}}]\!],\ 
[\![\text{text\_quality},\ \text{reason},\ \boxed{\text{score}}]\!],\ 
[\![\text{context\_coherence},\ \text{reason},\ \boxed{\text{score}}]\!]

Where \texttt{score} is your numerical rating (0--3) and \texttt{reason} is your brief justification.

\textless \textbf{Annotation Documents}\textgreater

\vspace{1em}
\textbf{User Prompt:}

Now, please evaluate this turn based on the scoring criteria. Your score should be between 0 and 3. And output the result in the format:
\[
[\![\text{Evaluation Criterion}_1,\ \text{Reason},\ \boxed{\text{score1}}],\ [\text{Evaluation Criterion}_2,\ \text{Reason},\ \boxed{\text{score2}}],\ \ldots\!]
\]

\end{tcolorbox}
\end{minipage}
\caption{System and user prompts for local single-turn evaluation (score with reason).}
\label{tab:local_judge_prompt_reason}
\end{figure}

\begin{figure}[htbp]
\begin{minipage}{\textwidth}
\begin{tcolorbox}[width=\textwidth]

\textbf{System Prompt:}

You are a judge model for evaluating the overall quality in multi-turn visual dialogues. You will receive a conversation history, please read it carefully. Next, I will provide you with an evaluation list and corresponding scoring criteria. Please compare the two responses (ResponseA and ResponseB) and give your final preference. Your output needs to follow the format:

$
[\text{Evaluation Criterion}_1,\ \boxed{\text{ResponseA}}],\ 
[\text{Evaluation Criterion}_2,\ \boxed{\text{ResponseB}}],\ 
\ldots
$

\textless \textbf{Annotation Documents}\textgreater

\vspace{1em}
\textbf{User Prompt:}
Now, please evaluate the conversation history based on the scoring criteria. And output the result in the format:

$
[\text{Evaluation Criterion}_1,\ \boxed{\text{ResponseA}}],\ 
[\text{Evaluation Criterion}_2,\ \boxed{\text{ResponseB}}],\ 
\ldots,
$

\end{tcolorbox}
\end{minipage}
\caption{System and user prompts for global comparison evaluation (preference only).}
\label{tab:global_judge_prompt_direct_preference}
\end{figure}

\begin{figure}[htbp]
\begin{minipage}{\textwidth}
\begin{tcolorbox}[width=\textwidth]

\textbf{System Prompt:}

You are a scoring model for evaluating the overall quality in multi-turn visual dialogues. You will receive a conversation history, please read it carefully. Next, I will provide you with an evaluation list and corresponding scoring criteria. Please score the conversation history based on the scoring criteria and provide a reason for your score. Your output needs to follow the format:

\[
[\text{Evaluation Criterion}_1,\ \text{Reason},\ \boxed{\text{ResponseA}}],\ 
\ldots,\ 
\]

\textless \textbf{Annotation Documents}\textgreater

\vspace{1em}
\textbf{User Prompt:}

Now, please evaluate the conversation history based on the scoring criteria. And output the result in the format:

\[
[\text{Evaluation Criterion}_1,\ \text{Your Judge Reason},\ \boxed{\text{ResponseA}}],\ 
\ldots,
\]

\end{tcolorbox}
\end{minipage}
\caption{System and user prompts for global comparison evaluation (with reason).}
\label{tab:global_judge_prompt_reason_preference}
\end{figure}

\begin{figure}[htbp]
\begin{minipage}{\textwidth}
\begin{tcolorbox}[width=\textwidth]

\textbf{System Prompt:}

You are a judge model for evaluating the quality of a single turn in multi-turn visual dialogues. You will receive a conversation history, please read it carefully. Next, I will provide you with an evaluation list and corresponding scoring criteria. Please compare the two responses (ResponseA and ResponseB) and give your final preference. Your output must follow this exact format:

Evaluation list: [local\_image\_text\_consistency, perceptual\_quality, text\_quality, contextual\_coherence]

\[
[\text{local\_image\_text\_consistency},\ \boxed{\text{ResponseA}}],\ 
\ldots,
\]

Where "preference" is your preference between ResponseA and ResponseB.

\textless \textbf{Annotation Documents}\textgreater

\vspace{1em}
\textbf{User Prompt:}

Now, please evaluate this turn based on the scoring criteria. Your preference should be between 0 and 3. And output the result in the format:

\[
[\text{Evaluation Criterion}_1,\ \boxed{\text{ResponseA}}],\ 
\ldots,\ 
[\text{total\_preference},\ \boxed{\text{ResponseA}}]
\]

\end{tcolorbox}
\end{minipage}
\caption{System and user prompts for local turn evaluation without reasoning.}
\label{tab:local_judge_direct_preference}
\end{figure}

\begin{figure}[htbp]
\begin{minipage}{\textwidth}
\begin{tcolorbox}[width=\textwidth]

\textbf{System Prompt:}

You are a judge model for evaluating the quality of a single turn in multi-turn visual dialogues. You will receive a conversation history, please read it carefully. Next, I will provide you with an evaluation list and corresponding scoring criteria. Please compare the two responses (ResponseA and ResponseB) and give your final preference. Your output must follow this exact format:

Evaluation list: [local\_image\_text\_consistency, perceptual\_quality, text\_quality, contextual\_coherence]

\[
[\text{local\_image\_text\_consistency},\ \text{reason},\ \boxed{\text{ResponseA}}],\ 
\ldots,
\]

Where "preference" is your preference between ResponseA and ResponseB and "reason" is your brief justification.

\textless \textbf{Annotation Documents}\textgreater

\vspace{1em}
\textbf{User Prompt:}

Now, please evaluate this turn based on the scoring criteria. Your preference should be between 0 and 3. And output the result in the format:

\[
[\text{Evaluation Criterion}_1,\ \text{Reason},\ \boxed{\text{ResponseA}}]
\ldots,\ 
[\text{total\_preference},\ \boxed{\text{ResponseA}}]
\]

\end{tcolorbox}
\end{minipage}
\caption{System and user prompts for local turn evaluation with reasoning.}
\label{tab:local_judge_reason_preference}
\end{figure}

%

\begin{table}[ht]
\centering
\caption{Supplementary \textit{Score Evaluation} of \oureval{}: Human Agreement. Percentage of agreement between human annotators and different judge models. Each judgment is independently reviewed by multiple annotators.}
\resizebox{\textwidth}{!}{%
\small
\begin{tabular}{l|p{3.5cm}|ccccc|cccccc}
\toprule 
\multirow{3}{*}{\textbf{Settings}} & \multirow{3}{*}{\textbf{MLLMs}} & \multicolumn{5}{c|}{\textbf{Local Setting}} & \multicolumn{6}{c}{\textbf{Global Setting}} \\
\cmidrule(lr){3-7} \cmidrule(lr){8-13} 
& & \textbf{L1} & \textbf{L2} & \textbf{L3} & \textbf{L4} & \textbf{Avg.} & \textbf{G1} & \textbf{G2} & \textbf{G3} & \textbf{G4} & \textbf{G5} & \textbf{Avg.}\\
\midrule 
\multirow{12}{*}{\textbf{\begin{tabular}[l]{@{}c@{}}Score \\ Evaluation\end{tabular}}}
& Gemini-Flash* & 0.430 & 0.625 & 0.783 & 0.827 & 0.666 & 0.702 & 0.573 & 0.593 & 0.089 & 0.665 & 0.524 \\
& Gemini-Flash* (+reason) & \textbf{0.437} & \textbf{0.626} & 0.783 & \textbf{0.828} & \textbf{0.669} & 0.702 & 0.573 & 0.621 & \textbf{0.302} & 0.669 & \textbf{0.573} \\
& GPT-4.1 & 0.392 & \textbf{0.626} & 0.785 & 0.791 & 0.649 & 0.685 & 0.585 & 0.625 & 0.069 & 0.681 & 0.529 \\
& GPT-4.1 (+reason) & 0.401 & 0.626 & 0.787 & 0.786 & 0.650 & 0.706 & \textbf{0.597} & 0.613 & 0.060 & 0.681 & 0.531 \\
& GPT-4o & 0.400 & 0.558 & \textbf{0.791} & 0.807 & 0.639 & \textbf{0.710} & 0.577 & 0.625 & 0.052 & 0.681 & 0.529 \\
& GPT-4o (+reason) & 0.404 & 0.545 & \textbf{0.791} & 0.812 & 0.638 & 0.706 & 0.585 & \textbf{0.629} & 0.056 & 0.681 & 0.531 \\
& Gemini-Pro* & 0.401 & 0.588 & 0.777 & 0.705 & 0.618 & 0.664 & 0.587 & 0.555 & 0.150 & 0.660 & 0.523 \\
& Gemini-Pro* (+reason) & 0.408 & 0.598 & 0.783 & 0.705 & 0.623 & 0.709 & 0.559 & 0.623 & 0.105 & 0.636 & 0.526 \\
& Claude-thinking* & 0.406 & 0.614 & 0.738 & 0.674 & 0.608 & 0.686 & 0.556 & 0.619 & 0.180 & \textbf{0.686} & 0.546 \\
& Claude-thinking* (+reason) & 0.412 & 0.612 & 0.736 & 0.662 & 0.606 & 0.682 & 0.569 & 0.623 & 0.205 & 0.682 & 0.552 \\
& o4-mini & 0.429 & 0.621 & 0.774 & 0.714 & 0.634 & 0.627 & 0.525 & 0.598 & 0.108 & 0.672 & 0.506 \\
& o4-mini (+reason) & 0.428 & \textbf{0.626} & 0.781 & 0.715 & 0.638 & 0.675 & 0.549 & \textbf{0.638} & 0.128 & 0.675 & 0.533 \\
\bottomrule 
\end{tabular}
} 
\label{tab:human_agreement_scores} 
\end{table} 

\subsection{Details of \textit{Crucial Step Recognition} Evaluation}

\textit{Crucial Step Recognition} evaluates whether a model can accurately identify the user's underlying intent in multi-turn multimodal interactions—typically signaled by the initial seed question—and effectively track the evolving user needs and preferences throughout the dialogue. Moreover, it assesses whether the model can recognize which specific step fulfills the user's core intention. This capability is critical for enhancing task completion and user experience in human-AI interactions.

During evaluation, we collect human-annotated rationales for crucial step recognition as reference answers. Given recent findings that advanced models can achieve human-comparable performance in pairwise response comparison \citep{zheng2023judging, chen2024mllm}, we employ GPT-4o \citep{openai2024gpt4o} as the judge to perform partial order comparisons between model outputs (Figure \ref{tab:crucial_step_recognition} presents the system and user prompts for evaluated models) and the reference answers (Figure \ref{tab:judge_model_crucial_step_recognition} presents the system and user prompts for judge models). Each judgment is subsequently reviewed by three human experts, and only those achieving a predefined agreement threshold are considered valid. Final scores are computed by averaging across all evaluation points.

\begin{figure}[htbp]
\begin{minipage}{\textwidth}
\begin{tcolorbox}[width=\textwidth]

\textbf{System Prompt:}

You are a crucial step recognition model. You will receive a multi-turn dialogue. Based on the dialogue content, determine which steps are crucial and which are optional. Evaluate the model's performance in recognizing key steps and whether it completed the user's initial task.

\vspace{0.5em}
\textbf{Crucial Step Recognition:}

\textbf{Definition:} In multi-turn interactions, the model must accurately identify and complete crucial steps, avoiding irrelevant or incorrect information.

\vspace{0.5em}
\textbf{Example:}

\begin{itemize}
    \item \textbf{Task:} Step-by-step guidance for drawing a cat
    \item \textbf{Crucial steps:} Sketch outline $\rightarrow$ Refine facial features $\rightarrow$ Adjust proportions $\rightarrow$ Apply color
    \item \textbf{Incorrect:} Model asks user to color before the outline is drawn
    \item \textbf{Correct:} Model guides user through steps in a logical order
\end{itemize}

\vspace{1em}
\textbf{User Prompt:}

You are a crucial step recognition model. You will receive a multi-turn dialogue. Based on the dialogue content, determine which steps are crucial and which are optional. Evaluate the model's performance in recognizing key steps and whether it completed the user's initial task. Crucial Step Recognition \textbf{Definition}: In multi-turn interactions, the model must accurately identify and complete crucial steps, avoiding irrelevant or incorrect information. \textbf{Example}:

* \textbf{Task}: Step-by-step guidance for drawing a cat

* \textbf{Crucial steps}: Sketch outline $\rightarrow$ Refine facial features $\rightarrow$ Adjust proportions $\rightarrow$ Apply color

* \textbf{Incorrect}: Model asks the user to color before the outline is drawn

* \textbf{Correct}: Model guides the user through steps in a logical order

\end{tcolorbox}
\end{minipage}
\caption{System and user prompts for crucial step recognition in multi-turn dialogue.}
\label{tab:crucial_step_recognition}
\end{figure}

\begin{figure}[htbp]
\begin{minipage}{\textwidth}
\begin{tcolorbox}[ width=\textwidth]

\textbf{System Prompt:}

You are a \textbf{Judge Model} designed to evaluate a model's performance in identifying key steps within multi-turn dialogues. Your task is to compare two inputs:  
1. \textbf{Reference Answer}: The ideal, ground truth response from a model that accurately represents the correct interpretation of the dialogue.  
2. \textbf{Model Inference}: A model-generated response to the same multi-turn dialogue, which may differ from or match the \textbf{Reference Answer}.  

\vspace{0.5em}
\textbf{Scoring Criteria:}

\begin{itemize}
    \item \textbf{Score Range}: 1 to 5 (where 1 is the lowest, 5 is the highest).
    \item \textbf{How to Score}:
    \begin{itemize}
        \item \textbf{5}: Model Inference is flawless or better than Reference Answer. All key steps correct.
        \item \textbf{4}: Mostly correct with minor issues.
        \item \textbf{3}: Partially correct with significant omissions or errors.
        \item \textbf{2}: Many missing or wrong steps.
        \item \textbf{1}: Fundamentally incorrect or misinterprets dialogue.
    \end{itemize}
\end{itemize}

\vspace{0.5em}
\textbf{Evaluation Guidelines:}
\begin{itemize}
    \item Focus on key steps driving the dialogue.
    \item Evaluate clarity, accuracy, and logical flow.
    \item Determine if the Model Inference aligns with intended meaning.
\end{itemize}

\vspace{0.5em}
\textbf{Additional Notes:}
\begin{itemize}
    \item Note if Model Inference is better or worse than Reference Answer.
    \item Justify the score with detailed rationale.
    \item Provide recommendations or point out overlooked steps if necessary.
\end{itemize}

\vspace{1em}
\textbf{User Prompt:}

Now evaluate the following response and give your score and reason. Your score should be in the range of 1 to 5 and in the format of 'score: [[score]], reason: [[reason]]'. \{reference\_answer\} \{model\_inference\}"

\end{tcolorbox}
\end{minipage}
\caption{System and user prompts for evaluation and crucial step recognition.}
\label{tab:judge_model_crucial_step_recognition}
\end{figure}

\subsection{More Results}

\paragraph{A little knowledge is a dangerous thing} Table~\ref{tab:human_agreement_scores} reports the human agreement accuracy for \textit{Score Evaluation}. Notably, although models often assign identical scores to those given by human annotators, the resulting Pearson correlation coefficients are relatively low. This suggests that models may be guessing scores rather than capturing the nuanced distinctions in human ratings.

\paragraph{Pairwise comparison evaluated using spearman correlation}
\revision{We thank the reviewer HqHt for the careful reading, statistical insights, and constructive feedback. Inspired by the suggestion, we conduct a detailed pairwise comparison of model performance on \oureval{} using Spearman correlation. Table~\ref{tab:pairwise_spearman} summarizes the results.}

\begin{table}[h]
\centering
\caption{\oureval{} pairwise comparison results (metric: spearman).  L1--L4 and G1--G5 denote different local and global evaluation subsets, respectively.}
\label{tab:pairwise_spearman}
\resizebox{\textwidth}{!}{%
\begin{tabular}{lccccccccccc}
\toprule
Model & L1 & L2 & L3 & L4 & Avg. & G1 & G2 & G3 & G4 & G5 & Avg. \\
\midrule
GPT-4.1 & 0.1719 & 0.1365 & 0.1853 & 0.1562 & 0.1625 & 0.5592 & 0.5281 & 0.5446 & 0.4618 & 0.3111 & 0.4809 \\
GPT-4.1 (+reason) & 0.2175 & 0.1725 & 0.1482 & 0.1893 & 0.1819 & 0.3600 & 0.4543 & 0.4626 & 0.4710 & 0.0449 & 0.3586 \\
GPT-4o & 0.1192 & 0.0486 & 0.0889 & 0.1203 & 0.0942 & 0.3304 & 0.4550 & 0.3622 & 0.3193 & 0.2440 & 0.3422 \\
GPT-4o (+reason) & 0.0674 & 0.1014 & 0.0381 & 0.0864 & 0.0733 & 0.1341 & 0.2395 & 0.2472 & 0.2235 & 0.0655 & 0.1820 \\
Gemini-Pro* & 0.0787 & 0.0749 & 0.1354 & 0.0126 & 0.0754 & 0.5019 & 0.4662 & 0.2890 & 0.1052 & 0.0682 & 0.2861 \\
Gemini-Pro* (+reason) & 0.0535 & 0.0247 & 0.1055 & 0.0009 & 0.0461 & 0.4476 & 0.4188 & 0.4189 & 0.3228 & 0.1841 & 0.3585 \\
Claude-thinking* & 0.1562 & 0.2247 & 0.1345 & 0.1357 & 0.1628 & 0.5978 & 0.5614 & 0.5438 & 0.3720 & 0.1579 & 0.4466 \\
Claude-thinking* (+reason) & 0.1521 & 0.2007 & 0.1916 & 0.1585 & 0.1757 & 0.4779 & 0.4996 & 0.5364 & 0.4838 & 0.1421 & 0.4280 \\
o4-mini & 0.0291 & 0.1736 & 0.1311 & 0.1621 & 0.1240 & 0.5761 & 0.5310 & 0.5392 & 0.4415 & 0.2326 & 0.4641 \\
o4-mini (+reason) & 0.1234 & 0.1515 & 0.0962 & 0.1291 & 0.1250 & 0.4269 & 0.4725 & 0.5293 & 0.4998 & 0.1778 & 0.4213 \\
\bottomrule
\end{tabular}
}
\end{table}

\section{Experiment Details}
\label{app:experiment_details}

All training was conducted on 8 NVIDIA H800 GPUs. We used Qwen2.5-VL-3B-Instruct and Qwen2.5-VL-7B-Instruct as the backbone models for training the judge model. Table \ref{tab:training-config} presents the key training hyperparameters used in our experiments.

\begin{table}[h]
\centering
\caption{Key training hyperparameters used in our experiments.}
\small
\begin{tabular}{ll}
\toprule
\textbf{Parameter} & \textbf{Value} \\
\midrule
Number of GPUs & 8 $\times$ NVIDIA H800 \\
Epochs & 3 \\
Batch size (train/eval) & 8 / 8 \\
Gradient accumulation steps & 1 \\
Gradient checkpointing & True \\
Learning rate & 3e-5 \\
LR scheduler & constant\_with\_warmup \\
Warmup ratio & 0.03 \\
Adam betas & (0.9, 0.95) \\
Weight decay & 0.0 \\
Mixed precision & bf16=True, fp16=False \\
Evaluation strategy & epoch \\
Regularization coefficient & 0.001 \\
Freeze vision tower & True \\
Freeze language model & False \\
Freeze MM projection layer & False \\
Max token length & 8192 \\
\bottomrule
\end{tabular}
\label{tab:training-config}
\end{table}

\subsection{Preliminaries of Preference Modeling}

A widely adopted approach for modeling human preferences is to employ a preference predictor grounded in the Bradley–Terry (BT) model \citep{bradley1952rank}. Given a pair of answers $(\vy_1, \vy_2)$ generated from an question $\vx$, BT model indicates that the human preference distribution $p^{*}$ \citep{bai2022training, rafailov2024direct} can be expressed based on the underlying human reward function $r^{*}(\vy,\vx)$ as 
\begin{equation*}
    p^{*}(\vy_1 \succ \vy_2 | \vx) = \frac{\exp(r^{*}(\vy_1, \vx))}{\exp(r^{*}(\vy_1, \vx)) + \exp(r^{*}(\vy_2, \vx))},
\end{equation*}
where $\vx = (\vx_{\text{I}}, \vx_{\text{T}})$ and $\vy = (\vy_{\text{I}}, \vy_{\text{T}})$ represent the multimodal (image-text) input and output, respectively. Hence, given a human image-text preference dataset $\mathcal{D} = \{(\vx^{(i)},\vy_w^{(i)},\vy_l^{(i)})\}_{i=1}^N$, the training objective for a multimodal reward model $r_{\phi}(\vy,\vx)$ parameterized by $\phi$ is defined as:
\begin{equation*}
    \mathcal{L}(\phi,\mathcal{D}) = -\mathbb{E}_{(\vx,\vy_w,\vy_l) \sim \mathcal{D}}[\log\sigma(r_{\phi}(\vy_w,\vx)-r_{\phi}(\vy_l,\vx))]
\end{equation*}
However, when extending to multi-turn settings, new challenges arise—particularly in capturing the dynamics of evolving user preferences across turns. Moreover, traditional outcome-level reward signals often fail to generalize in purely textual domains \citep{shani2024multi}, let alone in complex multimodal settings involving interleaved understanding and generation. 
\ours{} incorporates both \textit{local} and \textit{global} human annotations in multi-turn, multimodal interactions, leading us to investigate efficient preference modeling methods under this more realistic and challenging scenario.

\subsection{Long Horizon Human Value Preference Modeling}

Inspired by \citep{qiu2024reward,liao2025tpo}, we investigate two strategies for modeling long-horizon preferences in multi-turn multimodal scenarios: \textit{prefix preference} and \textit{chain-based preference}. Let $\mathcal{D}_{\text{multi-turn}} = \{(\vx_1^{(i)},\vy_1^{(i)},\cdots, \vx_{k_i}^{(i)},\vy_{k_i}^{(i)})\}_{i=1}^N$ denote the multi-turn human image-text dataset, where $k_i$ denotes the number of turns for each conversation. The \textit{prefix-preference} approach models preferences at the \textit{turn level}. Given a prefix of the conversation history, it aims to identify the preferred candidate response for the current turn, effectively capturing fine-grained local preferences. The training objective for the \textit{prefix-preference} reward model $r_{\phi_{\text{prefix}}}(\vy,\vx)$ is 
\begin{equation*}
    \mathcal{L}(\phi_{\text{prefix}},\mathcal{D}_{\text{prefix}}) = - \mathbb{E}_{(\vz,\vy_k^w,\vy_k^l)\sim\mathcal{D}_{\text{prefix}}}[\log\sigma(r_{\phi_{\text{prefix}}}(\vy_k^w,\vz)-r_{\phi_{\text{prefix}}}(\vy_k^l,\vz)],
\end{equation*}
where $\vz = (\vx_1,\vy_1,\cdots,\vx_k)$ stands for the shared prefix of the different conversations, and the \textit{prefix-preference} dataset is denoted as $\mathcal{D}_{\text{prefix}} = \{(\vz,\vy_{k}^w,\vy_{k}^l)|(\vz,\vy_{k}^w), (\vz,\vy_{k}^l)\sim \mathcal{D}_{\text{multi-turn}}\}$.

In contrast, the \textit{chain-based preference} approach models preferences at the \textit{conversation level} by comparing complete conversation trajectories conditioned on the same \textit{seed question} $\vx_1$. It seeks to capture the human’s overall intent and preference across the entire multi-turn dialogue. The training objective for the \textit{chain-based preference} reward model $r_{\phi_{\text{chain}}}(\vy,\vx)$ is defined as,
\begin{equation*}
    \mathcal{L}(\phi_{\text{chain}},\mathcal{D}_{\text{chain}}) = - \mathbb{E}_{(\vx_1,\vw^w,\vw^l) \sim \mathcal{D}_{\text{chain}}}[\log\sigma(r_{\phi_{\text{chain}}}(\vw^w,\vx_1)-r_{\phi_{\text{chain}}}(\vw^l,\vx_1))],
\end{equation*}
where $\vw = (\vy_0,\vx_1,\vy_1, \cdots, \vx_k,\vy_k)$ represents the whole conversation chain and the \textit{chain-based} preference dataset is $\mathcal{D}_{\text{chain}} = \{(\vx_0,\vw^w,\vw^l)|\vy_0^w\neq\vy_0^l \bigwedge
 (\vx_0,\vw^w), (\vx_0,\vw^l)\sim \mathcal{D}_{\text{multi-turn}}\}$.

\subsection{Evaluation Details}
Due to the absence of publicly available human-annotated test sets for multi-turn multimodal interactions, we adopt a random 9:1 train-test split strategy, ensuring that no \textit{seed question}s appears in both sets. To investigate the multi-turn scaling law of judge models, we ensure that the compared groups with different numbers of communication turns are matched in both data volume and computational cost. We repeat experiments at varying data scales to validate the robustness of our conclusions.

\section{Application}
\subsection{Application in Post-training Methods}
\label{sec:post_training}

\begin{table}[h!]
  \centering
  \caption{InterMT-enhanced alignment improves single-turn and multi-turn multimodal understanding and generation.}
  \label{tab:main_results}
  \resizebox{\textwidth}{!}{%
  \begin{tabular}{l|cccc|ccc}
    \toprule
    \textbf{Task} & \multicolumn{4}{c|}{\textbf{MMIE (Single-turn)}} & \multicolumn{3}{c}{\textbf{Interactor (Ours, Multi-turn)}} \\
    \midrule
    \textbf{Models} & \textbf{Situational Analysis} & \textbf{Project-based Learning} & \textbf{Multi-step Reasoning} & \textbf{Weighted Average} & \textbf{Local-Preference} & \textbf{Global-Preference} & \textbf{Average} \\
    \midrule
    \multicolumn{8}{c}{\textit{Closed-source Models}} \\
    \midrule
    GPT4o+SD-XL & 59.2\% & 68.6\% & 49.7\% & 59.8\% & 77.3\% & 56.3\% & 66.8\% \\
    GPT4o+SD-3 & 58.4\% & 68.7\% & 48.9\% & 59.3\% & 75.5\% & 54.8\% & 65.2\% \\
    GPT4o+Flux & 58.2\% & 69.2\% & 48.7\% & 59.4\% & 77.7\% & 54.5\% & 66.1\% \\
    Claude+SD-XL & 53.9\% & 68.6\% & 48.3\% & 57.6\% & 77.2\% & 61.0\% & 69.3\% \\
    Claude+SD-3 & 53.5\% & 68.6\% & 48.5\% & 57.5\% & 72.8\% & 59.2\% & 66.0\% \\
    Claude+Flux & 53.3\% & 68.8\% & 48.8\% & 57.6\% & 74.5\% & 60.7\% & 67.6\% \\
    \midrule
    \multicolumn{8}{c}{\textit{Open-source Models}} \\
    \midrule
    AA-Chameleon & 41.8\% & 46.3\% & 59.5\% & 44.7\% & 18.2\% & 16.0\% & 17.1\% \\
    AA-Chameleon+RS & 43.4\% & 45.3\% & 60.3\% & 45.4\% (+0.7\%) & 18.7\% & 17.8\% & 18.3\% (+1.2\%) \\
    AA-Chameleon+SFT & \textbf{46.3\%} & \textbf{56.9\%} & \textbf{65.6\%} & \textbf{56.6\% (+11.9\%)} & \textbf{22.5\%} & \textbf{21.3\%} & \textbf{21.9\% (+4.8\%)} \\
    AA-Chameleon+DPO & 45.8\% & 58.5\% & 59.0\% & 54.4\% (+9.7\%) & 20.7\% & 18.5\% & 19.6\% (+2.5\%) \\
    \midrule
    Janus-Pro & 60.3\% & 55.3\% & 43.4\% & 53.0\% & 19.5\% & 18.0\% & 18.8\% \\
    Janus-Pro+RS & \textbf{61.6\%} & 55.1\% & 42.2\% & 53.9\% (+0.9\%) & \textbf{25.7\%} & \textbf{22.0\%} & \textbf{23.8\% (+5.0\%)} \\
    Janus-Pro+SFT & 50.2\% & 57.3\% & \textbf{58.7\%} & \textbf{55.4\% (+2.4\%)} & 21.0\% & 18.5\% & 19.8\% (+1.0\%) \\
    Janus-Pro+DPO & 50.5\% & \textbf{63.0\%} & 50.9\% & 54.8\% (+1.8\%) & 20.2\% & 19.5\% & 19.8\% (+1.0\%) \\
    \bottomrule
    
  \end{tabular}%
  }
\end{table}

\revision{Our \texttt{InterMT} dataset is designed to be a versatile resource for enhancing Large Vision-Language Models (LVLMs) through various post-training alignment techniques. To demonstrate its efficacy, we applied \texttt{InterMT} to improve two strong open-source models, AA-Chameleon \citep{ji2024align} and Janus-Pro \citep{chen2025janus}, using three distinct alignment methods: Supervised Fine-Tuning (SFT) \citep{ouyang2022training}, Rejection Sampling (RS) \citep{touvron2023llama}, and Direct Preference Optimization (DPO) \citep{rafailov2024direct}. The results, summarized in Table~\ref{tab:main_results}, show that fine-tuning with \texttt{InterMT} improves performance across both single-turn vision-language benchmarks (\textit{e.g.}, MMIE \citep{xia2025mmie}) and complex multi-turn multimodal tasks (\textit{e.g.}, our Interactor benchmark\footnote{Interactor is an evaluation set derived from real human data, independent of the training dataset}).}

We train the models based on the \texttt{Align-Anything} codebase\footnote{\url{https://github.com/PKU-Alignment/align-anything}}, and implement MMIE following the original benchmark protocols. For each sub-metric, evaluation scores are computed as a weighted average according to the number of valid evaluation instances, consistent with the scoring methodology of the original MMIE benchmark.

It is worth noting that, due to the current absence of fully native multimodal input-output foundation models—most existing models support multiple modes such as TI2T, I2T, and T2I, but can activate only a single mode at a time—InterMT may not yet realize its full potential in a single training pass. During fine-tuning, we need to select which mode first to activate to most effectively enhance performance. Nevertheless, as demonstrated by both \oureval{} ~~and the publicly released \texttt{InterMT-Judge}, even in the absence of fully multimodal native models, \ours{} enables promising explorations in preference modeling and lays the groundwork for future advances in multi-turn multimodal human preference alignment.
\end{document}